\documentclass[11pt]{article}
\usepackage{bbm}

\usepackage{epsfig,endnotes}
\usepackage{multirow}
\usepackage{subfig}
\usepackage{graphicx}
\usepackage{grffile}

\newcounter{subcopyrightbox@save}
\usepackage[font=bf]{caption}
\usepackage{color, url}
\usepackage{xspace} 

\usepackage{amsmath}

\usepackage{mathrsfs}
\usepackage{amssymb}
\usepackage{amsmath}
\usepackage{amsthm}
\usepackage{epstopdf}
\usepackage{balance}
\usepackage{bm}

\usepackage{thm-restate,thmtools}

\newtheorem{definition}{Definition}

\usepackage[ruled,linesnumbered,noend]{algorithm2e}
\SetKwRepeat{Do}{do}{while}
\usepackage{mathtools}

\usepackage{cleveref}
\usepackage{geometry}
\usepackage{changepage}

\newcommand{\myparatight}[1]{\smallskip\noindent{\bf {#1}:}~}

\newcommand{\RN}[1]{%
  \textup{\uppercase\expandafter{\romannumeral#1}}%
}

\AtBeginDocument{%
  \providecommand\BibTeX{{%
    \normalfont B\kern-0.5em{\scshape i\kern-0.25em b}\kern-0.8em\TeX}}}

\begin{document}

\title{\LARGE{\bf{GraphFL: A Federated Learning Framework for Semi-Supervised Node Classification on Graphs}}}

 \author{Binghui Wang, Ang Li, Hai Li, and Yiran Chen \\ ECE Department, Duke University \\ \{binghui.wang,  ang.li630, hai.li, yiran.chen\}@duke.edu } 
          
   \date{}

   \maketitle

\begin{abstract}
\label{abstract}
Graph-based semi-supervised node classification 
(GraphSSC) 
has wide applications, ranging from networking and security to data mining and machine learning, etc. 
However, existing centralized GraphSSC methods are impractical to solve many real-world graph-based problems, as collecting the entire graph and labeling a reasonable number of labels is 
time-consuming and costly, and data privacy may be also violated. 
Federated learning (FL) is an emerging learning paradigm that enables collaborative learning among multiple clients, which can mitigate the issue of label scarcity and protect data privacy as well. 
Therefore, performing GraphSSC under the FL setting is a promising solution to solve real-world graph-based problems.
However, existing FL methods 1) perform poorly when data across clients are non-IID, 2) cannot handle data with new label domains, and 3) cannot leverage unlabeled data, while all these issues naturally happen in real-world graph-based problems. 

To address the above issues, we propose the first FL framework, namely GraphFL, for semi-supervised node classification on graphs. Our framework is motivated by meta-learning methods. 
Specifically, we propose two GraphFL methods to respectively address the
non-IID issue in graph data and handle the tasks with new label domains.   
Furthermore, we design a self-training method to leverage unlabeled graph data.
We adopt representative graph neural networks as GraphSSC methods and evaluate GraphFL on multiple graph datasets. Experimental results demonstrate that GraphFL significantly outperforms the compared FL baseline and GraphFL with self-training can obtain better performance.    

\end{abstract}
\section{Introduction}
\label{intro}

Given a graph and a small number of labeled training nodes, GraphSSC (GraphSSC) is to predict the labels of testing nodes in the graph.  
GraphSSC has various applications such as graph-based fraud detection~\cite{Yu06,Yu08,sybilrank,sybilbelief,
wang2017sybilscar, jia2017random, wang2017gang,wang2011review,akoglu2013opinion,li2014spotting,rayana2015collective,weber2019anti}, graph-based attribute inference~\cite{jia2017attriinfer,gong2018attribute},
graph-based document classification~\cite{kipf2017semi,sen2008collective}, to name a few. 
However, existing GraphSSC methods are performed in a centralized learning manner, making them impractical to solve many real-world graph-based problems. 
For instance, consider the problem of detecting fake users in social networks (e.g., Facebook). 
It is time-consuming and 
costly to collect the entire social graph and obtain a reasonable amount of labeled nodes by a single party to perform centralized GraphSSC. When only a partial graph or/and limited number of labels are available, existing GraphSSC
methods achieve performance far from satisfactory (See Figures~\ref{impact_nodes_GCN} and~\ref{impact_nodes_SGC}). 
Moreover, centralized learning methods need to access the raw data and could violate data privacy.

Federated learning (FL)~\cite{mcmahan2016communication} was a recently proposed technique that enables collaborative learning among multiple parties/clients. It aims to mitigate the issue of data/label scarcity and protect data privacy as well. 
In the FL setting, there are multiple clients and a server, where each client is assumed to have limited labeled data and uses them to train a local model; the server learns a global model for prediction 
by aggregating the local client models in a privacy-preserving manner. 
Therefore, incorporating FL into GraphSSC is a promising solution to solve real-world large graph-based problems. 
In this context, each client may have a partial graph of the original large graph and has a very few labeled nodes. 
For instance, in fake user detection on Facebook, each client could be a mobile user who may only collect an ego-network and the labeled users are his verified Facebook accounts in his ego-network. However, we note that 
directly applying existing FL methods to solve GraphSSC problems faces the following challenges. 
1) Existing FL methods perform poorly when data across clients are non-independent identically distributed (non-IID)~\cite{mcmahan2016communication,zhao2018federated,li2019federated}. However, graph data are essentially and potentially highly non-IID across clients. 
For instance, sampling and labeling representative nodes from a large graph is challenging~\cite{leskovec2006sampling}, and the limited labeled nodes in different clients could hardly have the same distribution. 
2) Existing FL methods focus on the problems where training data and testing data share the same label domain. However, real-world graphs are dynamically changing and new types of testing nodes can emerge at any time. For instance, new-type accounts could appear anytime in social networks. 
3) Existing FL methods are mainly for supervised learning and cannot leverage unlabeled data. However, 
real-world GraphSSC problems often involve limited labeled nodes, while having a substantial number of unlabeled nodes.

\myparatight{Our work} We design a novel FL framework, namely, GraphFL, to perform graph-based semi-supervised node classification and address the above challenges. To the best of our knowledge, this is the first work targeting GraphSSC problems under the FL setting. 
Our framework is motivated by a recent meta-learning method 
called model-agnostic meta-learning (MAML)~\cite{finn2017model}, which demonstrates fast adaptation ability to new tasks. 
Given a set of tasks drawn from an underlying distribution, MAML learns a task-independent initialization that performs well on all tasks after a few steps of gradient updates. 

We observe that MAML is apt for the FL setting. Specifically, we can treat each task as a client and the task-independent initialization as a global model learnt on the server.
Inspired by this observation, we propose two GraphFL methods aiming to address Challenge 1 and Challenge 2, respectively. 
To address Challenge 1, we note that data across tasks do not require to be IID in MAML and thus propose to incorporate MAML into FL. Specifically, our method has two stages. 
First, we learn a global model on the server by following the training scheme of MAML and thus can mitigate the issue caused by non-IID graph data. Then, we leverage existing FL methods to further update the global model such that it can achieve good generalization on testing nodes.
To address Challenge 2, we propose to reformulate MAML in the FL framework, where we define a novel objective function different from the existing FL methods. In doing so, we can learn a shared global model on the server for all clients, such that the global model can fast adapt to testing nodes whose label domains are different from training nodes'. 
 We further propose a self-training method to address Challenge 3.
Specifically, we first train a local model in each client using a GraphSSC method with the client's labels.  
We then use each trained local model to predict the client's unlabeled nodes and select 
the unlabeled nodes with the most confident predictions. 
These selected nodes as well as their predicted labels are used as extra labeled nodes to train our GraphSSC methods. 

We adopt two representative graph neural networks, i.e., Graph Convolutional Network (GCN)~\cite{kipf2017semi}
and Simple Graph Convolution (SGC)~\cite{wu2019simplifying}, as the GraphSSC methods, and incorporate our GraphFL into GCN and SGC to perform federated semi-supervised node classification.
We evaluate GraphFL on several benchmark graph datasets. Our results demonstrate that GraphFL significantly outperforms the compared FL baseline when the labeled nodes across clients are highly non-IID; GraphFL shows a better ability to handle testing nodes with new label domains than FL; and GraphFL with self-training can further obtain better performance.    
Our contributions can be summarized as follows:

\begin{itemize}
\item We propose GraphFL, the first federated semi-supervised node classification method on graphs.

\item GraphFL addresses the non-IID issue in graph data; handles testing nodes with new label domains; and leverages unlabeled nodes via self-training.

\item We evaluate GraphFL for federated GraphSSC on multiple graph datasets; and show the superiority of GraphFL over compared FL baseline.  

\end{itemize}

\section{Related Work}
\label{related}

\myparatight{Semi-supervised node classification on graphs}
Graph-based semi-supervised node classification has been widely studied and various methods have been proposed from multiple research fields including networking~\cite{Yu06,sybilrank,wang2017sybilscar,wang2018structure}, security~\cite{Yu08,sybilbelief,gong2016you,jia2017random,wang2019graph}, data mining~\cite{yang2012analyzing,akoglu2013opinion,li2014spotting,rayana2015collective,pham2016anomaly,jia2017attriinfer,wang2017gang,weber2019anti}, machine learning~\cite{zhu2003semi,yang2016revisiting,kipf2017semi,velivckovic2017graph,wu2019simplifying}, etc. 
Conventional methods include label propagation~\cite{zhu2003semi,zhou2003learning}, iterative classification~\cite{lu2003link}, manifold regularization~\cite{belkin2006manifold}, belief propagation~\cite{gatterbauer2015linearized}, to name a few. 
Graph neural networks (GNNs)~\cite{kipf2017semi,velivckovic2017graph,hamilton2017inductive,xu2018representation,wu2019simplifying} are recent methods which generalize neural networks to graph data for semi-supervised node classification. 
For instance, 
GCN~\cite{kipf2017semi} and SGC
are two representative GNNs. GCN stacks layers of learnable first-order spectral filters, which are motivated by spectral graph convolutions~\cite{duvenaud2015convolutional}, followed by a nonlinear activation function.  
SGC is a variant of GCN that removes nonlinear activation functions between GCN layers.
SGC is computationally more efficient than GCN and its performance is comparable to GCN.
GNNs have demonstrated to outperform conventional methods for semi-supervised node classification.
In this paper, we focus on using GNNs as the semi-supervised node classification methods for simplicity. 

\myparatight{Federated learning (FL)}
FL~\cite{mcmahan2016communication,smith2017federated,zhao2018federated,li2019federated,li2020convergence} is an emerging
distributed learning paradigm which can collaboratively train multiple client models and maintain a shared global model on a server. 
FL can mitigate the issue of data/label scarcity and protect clients' data privacy as well. 
Specifically, each client is assumed to have very limited labeled data and trains a local model using its labeled data. The server learns the global model by aggregating the local client models in a privacy-preserving manner. 
For instance, FedAvg~\cite{mcmahan2016communication}, the most widely used FL method, adopts averaging to aggregate the local models. 
The purpose of FL is to learn a global model that achieves good performance for all clients. 
However, as shown in~\cite{zhao2018federated,li2019federated,li2020convergence}, when the data across the clients are non-IID, the existing FL methods fail to learn a global model that has a good generalization ability. 
Besides the non-IID issue, the existing FL methods cannot handle new data whose label domains are different from the training data' and they cannot leverage unlabeled data.   
However, we note that all these issues exist in real-world graph-based semi-supervised node classification problems.


\section{Problem Definition and Background}
\label{problem}

\subsection{Problem definition}
Suppose we are given a set of $I$ clients $\mathbb{C} = \{C^{(1)}, C^{(2)}, \cdots, C^{(I)} \}$, where each client $C^{(i)}$
owns a graph
\footnote{We interchangeably use client and client graph in this paper.}   
$G^{(i)} = (\mathbb{V}^{(i)}, \mathbb{E}^{(i)})$
with $\mathbb{V}^{(i)}$ the node set and $\mathbb{E}^{(i)}$ the edge set. 
Each node $v^{(i)} \in \mathbb{V}^{(i)} $ is associated with a feature vector $\mathbf{x}_{v^{(i)}}$ and a label $y_{v^{(i)}}$ among the label set $\mathbb{K} = \{1, 2, \cdots, K\}$. 
Moreover, each client graph $G^{(i)}$ has a set of a few labeled nodes $\mathbb{L}^{(i)} \subset \mathbb{V}^{(i)}$.
We assume each client $C^{(i)}$ can learn a local semi-supervised node classification model $f_{\theta^{(i)}}$, parameterized by $\theta^{(i)}$, based on its labeled set $\mathbb{L}^{(i)}$ and graph $G^{(i)}$. 
We also consider a server $\mathbb{S}$ that can learn global model parameters $\theta$ by aggregating the clients' local model parameters $\{ \theta^{(i)} \}_{i=1}^I$,
while not accessing the client graphs. 
Now, suppose we have a set of testing nodes $\mathbb{T}$ which may or may not have the same label domain as the training nodes.
Then, our problem 
of federated semi-supervised node classification on graphs 
is defined as follows:

\begin{definition}
Given a set of $I$ client graphs $\{G^{(i)}\}_{i=1}^I $ with labeled nodes $\{\mathbb{L}^{(i)}\}_{i=1}^I$, 
a server $\mathbb{S}$, and a set of testing nodes $\mathbb{T}$, our goal is to predict the testing nodes' labels
based on the global model parameters $\theta$ learnt on the server, which is the aggregation of local model parameters $\{\theta_i\}_{i=1}^I$ learnt on the clients. 
\end{definition}

\myparatight{Design goals}
We aim to design a federated GraphSSC method that can achieve the  
following three goals: 
1) addressing the non-IID issue in graph data; and 2) generalizing to testing nodes with new label domains; and 3) leveraging the unlabeled nodes in clients.  
Next, we introduce model-agnostic meta-learning (MAML)~\cite{finn2017model}, which inspires the design of our method. 

\subsection{Model-agnostic meta learning (MAML)}
Given a set of training tasks $\{{T}_i\}$ drawn from an underlying task distribution $\mathcal{T}$, instead of learning a model that performs well on all tasks, 
MAML~\cite{finn2017model} learns a task-independent initialization $\theta$ that performs well on all tasks after a few steps of gradient updates. 
Specifically, each task $T_i$ splits its labeled training set $\mathbb{L}^{(i)}$ into 
 a support set $\mathbb{L}_{S}^{(i)}$ and a disjoint query set $\mathbb{L}_{Q}^{(i)}$.
Then, MAML has a two-level optimization: inner-optimization and meta-optimization.  
In the inner-optimization, for each task ${T}_i$, MAML trains a model $f_{\theta}$ on 
the support set $\mathbb{L}_{S}^{(i)}$ with the initialization $\theta$ and outputs a task-specific model parameters $\theta^{(i)}$. 
MAML then takes each $\theta^{(i)}$ as the initialization and evaluates the model $f_{\theta^{(i)}}$ on the corresponding 
query set $\mathbb{L}_{Q}^{(i)}$ with a task-specific loss. 
In the meta-optimization, MAML minimizes the total loss on the query sets of all tasks 
simultaneously to learn the task-independent initialization.  
Formally, the objective function of MAML is as follows:
{
\begin{equation}
\min_{\theta} \mathcal{L}(\theta) = \sum_{T_i \sim \mathcal{T}} \mathcal{L}_{\mathbb{L}_{Q}^{(i)}} (\theta^{(i)}) = \sum_{T_i \sim \mathcal{T}} \mathcal{L}_{\mathbb{L}_{Q}^{(i)}} (\theta - \alpha \cdot \nabla \mathcal{L}_{\mathbb{L}_{S}^{(i)}}(\theta)),
\label{meta-opt}
\end{equation}
}%
where we use one step of gradient descent in the inner-optimization for simplicity; $\alpha$ is the learning rate;
 and the task-specific loss over the support set and the query set are respectively defined as
 {
 \begin{align}
&\mathcal{L}_{\mathbb{L}_{S}^{(i)}}(\theta) = \frac{1}{|\mathbb{L}_{S}^{(i)}|} \sum_{(x, y) \in \mathbb{L}_{S}^{(i)}} \ell(f_\theta(x), y), \\
&\mathcal{L}_{\mathbb{L}_{Q}^{(i)}}(\theta^{(i)}) = \frac{1}{|\mathbb{L}_{Q}^{(i)}|} \sum_{(x, y) \in \mathbb{L}_{Q}^{(i)}} \ell(f_{\theta^{(i)}}(x), y),
\end{align}
}%
where $\ell(\cdot)$ is the loss function defined for the specific task. 

The objective function of MAML in Equation~\ref{meta-opt} is solved via gradient descent with a meta-learning rate $\beta$, i.e.,
\begin{align}
\theta \leftarrow \theta - \beta \cdot \nabla \mathcal{L}(\theta).
\end{align}
MAML proceeds in an episodic manner, where in each episode a batch of tasks are sampled from the task distribution $\mathcal{T}$ for training. 
When a new task arrives, MAML uses the learnt task-independent initialization as the initial model and updates the model via a few steps of gradient descent with respect to the loss defined in  the new task. Then, the updated model is used to make predictions. 

\myparatight{Connection with FL} We observe that MAML is apt for the FL setting.
Specifically, if we treat each task as a client and the task-independent initialization as a global model learnt on the server, then MAML naturally fits the FL. 
Inspired by this observation, we incorporate MAML into FL and propose a GraphFL framework to study graph-based semi-supervised node classification problems. 
We note that~\cite{chen2018federated,fallah2020personalized} also have similar observations.

\section{The Proposed GraphFL Framework}
\label{alg}

We propose a novel FL framework for semi-supervised node classification on graphs and aim to achieve the above goals. 
Our framework mainly incorporates MAML into FL, and we name our framework as GraphFL.  
First, we develop two GraphFL methods that aim to address the non-IID issue in graph data and handle the testing nodes with new label domains, respectively.
Then, we design a self-training method to leverage unlabeled nodes in client graphs.

\subsection{GraphFL for federated GraphSSC with non-IID graph data}
\label{addr_nonIID}
One key challenge to solving federated GraphSSC is that graph data are essentially and potentially highly non-IID across clients, while existing FL methods perform poorly with non-IID data~\cite{mcmahan2016communication,zhao2018federated,li2019federated}. 
This is because the goal of existing FL is to collaboratively learn a global model for all clients, but 
the learnt global model cannot generalize well on clients' data when they are non-IID. 
We design a novel FL method GraphFL for federated GraphSSC that can handle the non-IID graph data across clients. 
Our GraphFL method is inspired by MAML, as it fits into the FL framework and trains a set of tasks simultaneously but does not require the data to be IID across the tasks.  
Specifically, our GraphFL method incorporates MAML into FL and learns a global model on the server that shows good generalization on testing nodes.  
GraphFL consists of two stages: {\bf Stage I} learns a global model on the server by following the training scheme of MAML and thus can mitigate the issue caused by non-IID graph data. {\bf Stage II} leverages the existing FL method to further update the global model such that it can achieve a good generalization ability.

\begin{algorithm}[t]

	\caption{GraphFL: non-IID graph data}
	\label{algo:GraphSSC_share}
	\LinesNumbered 
    \begin{flushleft}
	\textbf{Input:} Client graphs $\{G^{(i)}\}_{i}$, support nodes $\{\mathbb{L}_{S}^{(i)}\}_{i}$ and query nodes $\{\mathbb{L}_{Q}^{(i)}\}_i$, initial global model $\theta_0$ on the server, learning rate $\alpha$, meta-learning rate $\beta$, \#episodes $T$, \#epochs $T_e$, fraction $\rho$ of participating clients.
	Testing nodes $\mathbb{T}$.
	 
	// {\bf Training} \\
	\For {episode $t=0,1,...,T-1$}{
		
		Server randomly samples a set $\mathbb{C}_t$ of $\rho I$ clients \\
		Server sends the initialized global model $\theta_t$ to clients $\mathbb{C}_t$ \\
		{\bf {\bf Stage I}: MAML} \\
		// {\bf Client update} \\
		\For {\emph{each client $C^{(i)} \in \mathbb{C}_t$}}{
			
			$ \theta_t^{(i)}  \leftarrow \theta_t $
		
			\For {epoch $t=1,2,...,T_e$} { 
				\quad $\theta_t^{(i)} \leftarrow \theta_t^{(i)} - \alpha \cdot \nabla \mathcal{L}_{\mathbb{L}_{S}^{(i)}}(\theta_t^{(i)})$ 
			}

			$g_i \leftarrow \nabla_{\theta} \mathcal{L}_{\mathbb{L}_{Q}^{(i)}}(\theta_t^{(i)})$  
		
		}

		// {\bf Server update} \\
		$\hat{\theta}_t \leftarrow \theta_t - {\beta} \cdot \sum_{i \in \mathbb{C}_t} g_i$ \\
		{\bf {\bf Stage II}: FL} \\	
		// {\bf Client finetuning} \\
		\For {\emph{each client $C^{(i)} \in \mathbb{C}_t$}}{
			$ \hat{\theta}_t^{(i)}  \leftarrow \hat{\theta}_t $

			\For {\emph{each epoch $t=1,2,...,T_e$}} { 

				\quad $\hat{\theta}_t^{(i)} \leftarrow \hat{\theta}_t^{(i)} - \alpha \cdot \nabla \mathcal{L}_{\mathbb{L}_{S}^{(i)}}(\hat{\theta}_t^{(i)})$ 
			}
		}

		// {\bf Server aggregation (FedAvg)} \\
		$\theta_{t+1} \leftarrow \frac{1}{|\mathbb{C}_t|} \sum_{C^{(i)} \in \mathbb{C}_t} \hat{\theta}_t^{(i)} $
	}	
	
	// {\bf Testing} \\
	Predict labels of testing nodes $\mathbb{T}$ using the global model $\theta_T$
	\end{flushleft}
	
\end{algorithm}

\begin{algorithm}[t]
	\SetAlgoLined
	\caption{GraphFL: new label domain}
	\label{algo:GraphSSC_cross}
	\begin{flushleft}
	\textbf{Input:} Client graphs $\{G^{(i)}\}_{i}$, support nodes $\{\mathbb{L}_{S}^{(i)}\}_{i}$ and query nodes $\{\mathbb{L}_{Q}^{(i)}\}_i$, initial global model $\theta_0$ on the server, learning rate $\alpha$, meta-learning rate $\beta$, \#episodes $T$, \#epochs $T_e$, fraction $\rho$ of participating clients.
	Testing nodes $\mathbb{T}$.
	
	// {\bf Training} \\
	\For {episode $t=0,1,...,T-1$}{
		
		Server randomly samples a set $\mathbb{C}_t$ of $\rho I$ clients \\
		Server sends the global model $\theta_t$ to clients $\mathbb{C}_t$
		
		// {\bf Client update} \\
		\For {\emph{each client $C^{(i)} \in \mathbb{C}_t$}}{
			
			$ \theta_t^{(i)}  \leftarrow \theta_t $

			\For {epoch $t_e=1,2,...,T_e$} { 
				\quad $\hat{\theta}_t^{(i)} \leftarrow \theta_t^{(i)} - \alpha \cdot \nabla \mathcal{L}_{\mathbb{L}_{S}^{(i)}}(\theta_t^{(i)})$ 

				\quad $\theta_t^{(i)} \leftarrow \theta_t^{(i)} - \beta (\mathbb{I} - \alpha \nabla^2 \mathcal{L}_{\mathbb{L}_{S}^{(i)}}(\theta_t^{(i)})) \nabla \mathcal{L}_{\mathbb{L}_{Q}^{(i)}}(\hat{\theta}_t^{(i)})$ 
			}
		}
			
		// {\bf Server aggregation (FedAvg)}
		
		$\theta_{t+1} \leftarrow \frac{1}{|\mathbb{C}_t|} \sum_{C^{(i)} \in \mathbb{C}_t} \theta_t^{(i)} $
	
	}

	// {\bf Testing}
	
	Treat the global model $\theta_T$ as the initial model and update the model with a few labeled nodes from the new label domain

	Predict labels of testing nodes $\mathbb{T}$ using the updated model
    
    \end{flushleft}
    
\end{algorithm}

We first define some notations.
In each client $C^{(i)}$, we split the training set $\mathbb{L}^{(i)}$ into the support nodes $\mathbb{L}_{S}^{(i)}$ and the query nodes $\mathbb{L}_{Q}^{(i)}$. 
Suppose at episode $t$, the server $\mathbb{S}$ has global model parameters $\theta_t$, and each client $C^{(i)}$ has local model parameters $\theta_t^{(i)}$.
We define the loss on the support nodes $\mathbb{L}_{S}^{(i)}$ in client $C^{(i)}$ as 
${\mathcal{L}_{\mathbb{L}_{S}^{(i)}}(\theta_t) = \frac{1}{|\mathbb{L}_{S}^{(i)}|} \sum_{v^{(i)} \in \mathbb{L}_{S}^{(i)}} \ell(f_{\theta_t}(\mathbf{x}_v^{(i)}, G^{(i)}), y_v^{(i)})}$, and the loss on the query nodes $\mathbb{L}_{Q}^{(i)}$ as 
${\mathcal{L}_{\mathbb{L}_{Q}^{(i)}}(\theta_t^{(i)}) = \frac{1}{|\mathbb{L}_{Q}^{(i)}|} \sum_{v^{(i)} \in \mathbb{L}_{Q}^{(i)}} \ell(f_{\theta_t^{(i)}}(\mathbf{x}_v^{(i)}, G^{(i)}), y_v^{(i)})}$, where $f_{\theta_t}$ and $f_{\theta_t^{(i)}}$ are the GraphSSC models learnt on the support nodes and query nodes in client graph $G^{(i)}$, respectively.
Moreover, suppose at episode $t$, the server has learnt the global model parameters $\theta_t$. 
Then, our GraphFL method has the following steps.

\myparatight{Step I} The server randomly sends $\theta_t$ to a fraction $\rho$ of total clients, which we denote as $\mathbb{C}_t$. 

\myparatight{Step II} Each participating client $C^{(i)}$ in $\mathbb{C}_t$ first learns local model 
parameters $\theta_t^{(i)}$ by minimizing the loss on the support nodes $\mathbb{L}_{S}^{(i)}$ via gradient descent. Assuming one step of gradient descent, we have
{
\begin{equation}
\theta_t^{(i)} \leftarrow \theta_t - \alpha \cdot \nabla \mathcal{L}_{\mathbb{L}_{S}^{(i)}}(\theta_t), 
\end{equation}
}%
Next, each client $C^{(i)}$ evaluates the local model parameters $\theta_t^{(i)}$ on the query nodes $\mathbb{L}_{Q}^{(i)}$, obtains the gradient of the loss $\nabla_{\theta} \mathcal{L}_{\mathbb{L}_{Q}^{(i)}}(\theta_t^{(i)})$, and sends the gradient to the server.

\myparatight{Step III} The server leverages the gradients of all the participating clients $\mathbb{C}_t$ and updates the global model $\theta_t$ to be $\hat{\theta}_t$ via gradient descent. Assuming one step of gradient descent, we have
\begin{equation}
\hat{\theta}_t \leftarrow {\theta}_t - \beta \nabla_{\theta} \sum_{C^{(i)} \in \mathbb{C}_t} \mathcal{L}_{\mathbb{L}_{Q}^{(i)}}(\theta_t^{(i)}).
\end{equation}
With these steps by following the training scheme of MAML, the server learns a global model that can mitigate the non-IID issue in graph data. Next, we take the following two steps to further update the global model such that it can achieve a good generalization ability on all clients.

\myparatight{Step IV} Each participating client $C^{(i)}$ in $\mathbb{C}_t$ downloads the global model parameters $\hat{\theta}_t$ and finetunes the local model on the support nodes via gradient descent. Assuming one step of gradient descent, 
we have
\begin{equation}
\hat{\theta}_t^{(i)} \leftarrow \hat{\theta}_t - \alpha \cdot \nabla \mathcal{L}_{\mathbb{L}_{S}^{(i)}}(\hat{\theta}_t). 
\end{equation}

\myparatight{Step V} The server adopts the existing FL methods, e.g., FedAvg~\cite{mcmahan2016communication}, to update the global model. 
\begin{equation}
\label{server_update}
\theta_{t+1} \leftarrow \frac{1}{|\mathbb{C}_t|} \sum_{C^{(i)} \in \mathbb{C}_t} \hat{\theta}_t^{(i)}.
\end{equation}
After several episodes, the final global model 
is used to make predictions on testing nodes $\mathbb{T}$.
Algorithm~\ref{algo:GraphSSC_share} details our GraphFL method for federated GraphSSC with non-IID graph data.

\subsection{GraphFL for federated GraphSSC with new label domains}
\label{fl_new_domain}

Existing FL targets the problems where all data in clients share the same label domain. However, real-world graphs are dynamically changing and new types of nodes can emerge at any time. 
In this section, we study the graph-based problems where the training nodes and testing nodes have different label domains. One possible solution is leveraging \emph{transfer learning}. Specifically, we first learn a global model on the server based on the existing FL methods (e.g., FedAvg~\cite{mcmahan2016communication}). Next, we adopt the global model as the initial model and finetune the model using a few labeled nodes with new label domains. Then, we use the finetuned model to predict the labels of testing nodes with new label domains. 
However, as shown in our experimental results (See Table~\ref{impact_clients_cross}), such transfer learning-based solution achieves unsatisfactory performance. 

We design a novel GraphFL method for GraphSSC that can generalize to testing nodes with new label domains. 
Specifically, we propose to reformulate MAML in the FL framework and aim to learn a shared global model on the server for all clients, such that each client performs well after a few steps of gradient updates with respect to its loss defined by a specific GraphSSC method. 
Formally, we define our objective function as follows:
\begin{equation}
\label{obj_GraphSSC_cross}
\min_{\theta} \mathcal{L}(\theta) = \frac{1}{I} \sum_{i=1}^{I} \mathcal{L}_i(\theta)  = \frac{1}{I} \sum_{i=1}^{I} \mathcal{L}_{\mathbb{L}_{Q}^{(i)}} (\theta - \alpha \cdot \nabla \mathcal{L}_{\mathbb{L}_{S}^{(i)}}(\theta)), 
\end{equation}
where $\theta$ is the shared initialization we aim to learn; $\mathcal{L}_i(\theta)$ is the loss defined on 
client $C^{(i)}$. 
Note that our loss function is completely different from that in the existing FL methods~\cite{mcmahan2016communication,zhao2018federated,li2019federated}.

We now solve Equation~(\ref{obj_GraphSSC_cross}) in the FL framework.
Specifically, we first update the local model based on our defined client loss, and then update the global model by aggregating local models. Assume at episode $t$, server has global model parameters $\theta_t$. Then, GraphFL
has the following steps:

\myparatight{Step I} The server randomly sends $\theta_t$ to a fraction $\rho$ of total clients, which we denote as $\mathbb{C}_t$.

\myparatight{Step II} Each participating client $C^{(i)}$ in $\mathbb{C}_t$ learns local model parameters $\theta_t^{(i)}$ by minimizing its client loss $\mathcal{L}_i(\theta_t)$. Specifically, based on the global model parameters $\theta_t$, each client loss can be minimized via gradient descent. Assuming one step of gradient descent, we have:
\begin{equation}
\label{grad_des_GraphSSC_cross}
\theta_t^{(i)} \leftarrow \theta_t - \beta \cdot \nabla \mathcal{L}_i(\theta_t), 
\end{equation}
where $\nabla \mathcal{L}_i(\theta_t)$ is calculated as: 
{
\begin{align*}
\label{grad_GraphSSC_cross}
\nabla \mathcal{L}_i(\theta_t) = (\mathbb{I} - \alpha \cdot \nabla^2 \mathcal{L}_{\mathbb{L}_{S}^{(i)}}(\theta_t)) \cdot \nabla \mathcal{L}_{\mathbb{L}_{Q}^{(i)}} (\theta_t - \alpha \cdot \nabla \mathcal{L}_{\mathbb{L}_{S}^{(i)}}(\theta_t))).
\end{align*}
}%
Equation~(\ref{grad_des_GraphSSC_cross}) can be solved in two steps. 
First, the client $C^{(i)}$ obtains an intermediate model parameters $\hat{\theta}_t^{(i)}$ by running gradient descent with the loss defined on the labeled support nodes ${\mathbb{L}_{S}^{(i)}}$. Assuming one step of gradient descent, we have:
\begin{equation}
\hat{\theta}_t^{(i)} \leftarrow \theta_t - \alpha \cdot \mathcal{L}_{\mathbb{L}_{S}^{(i)}}(\theta_t).
\end{equation}
Next, each client $C^{(i)}$ updates its local model parameters $\theta_t^{(i)}$ using the labeled query nodes ${\mathbb{L}_{Q}^{(i)}}$. Formally, 
\begin{equation}
\theta_t^{(i)} \leftarrow \theta_t - \beta (\mathbb{I} - \alpha \nabla^2 \mathcal{L}_{\mathbb{L}_{S}}(\theta_t) \cdot \nabla \mathcal{L}_{\mathbb{L}_{Q}^{(i)}}(\hat{\theta}_t^{(i)}).
\end{equation}

\myparatight{Step III} The server updates the global model $\theta_{t+1}$ by minimizing the loss over participating clients $\mathbb{C}_t$ with gradient descent. I.e., 
\begin{equation}
\label{server_update}
\theta_{t+1} \leftarrow \theta_t - \frac{\beta}{|\mathbb{C}_t|} \sum_{C^{(i)} \in \mathbb{C}_t} \nabla \mathcal{L}_i(\theta_t) = \frac{1}{|\mathbb{C}_t|} \sum_{C^{(i)} \in \mathbb{C}_t} \theta_t^{(i)},
\end{equation}
where we substitute Equation~(\ref{grad_des_GraphSSC_cross}) in the last Equation. 
Equation~(\ref{server_update}) shows that the server updates the global model parameters by \emph{averaging} the local model parameters of participating clients, and this aggregation rule is exactly the same as FedAvg~\cite{mcmahan2016communication}.

By solving Equation~(\ref{obj_GraphSSC_cross}) with several episodes, we learn a global semi-supervised node classification model on the server that can fast adapt to testing nodes with new label domains. 
Specifically, we use the global model as the initial model, and update the model with a few steps of gradient descent using a few labeled nodes from new label domains. 
Then, we adopt the updated model to predict labels of testing nodes from new label domains. 
Algorithm~\ref{algo:GraphSSC_cross} details our GraphFL for federated GraphSSC with new label domains.

\subsection{Leveraging unlabeled nodes via self-training}
Existing FL methods are mainly for supervised learning, i.e., they only use labeled data.  
However, real-world graph-based problems often involve limited labeled nodes, while having a substantial number of unlabeled nodes. 
A natural question is how can we leverage unlabeled nodes to further enhance the performance of our GraphFL? 

We propose a self-training method to leverage unlabeled nodes in client graphs. 
Specifically, given a graph-based semi-supervised node classification method, we first use the method to train a local model in each client using the client's few labeled nodes. 
Next, in each client, we use its local model to predict unlabeled nodes and select a set of unlabeled nodes with 
the most confident predictions.
Then, we treat the predicted labels of the selected nodes as their pseudo labels, and 
add each client's selected nodes (as well as their pseudo labels) to the client's training set. 
Finally, we train our GraphFL methods on the augmented training sets for federated semi-supervised node classification. 
We observe in our experimental results (See Figure~\ref{impact_self_train}) that such a method is simple yet effective.

\section{Evaluation}
\label{eval}

\subsection{Experimental Setup}

\begin{table}[!t]\renewcommand{\arraystretch}{1.3}
\caption{Dataset statistics.}
\centering
\begin{tabular}{|c|c|c|c|c|} \hline 
{\small \textbf{Dataset}} & {\small \textbf{Cora}} & {\small \textbf{Citeseer}} & {\small \textbf{Coauthor CS}} & {\small \textbf{Amazon2M}} \\ \hline 
{\small \#Features} & {\small 1,433} & {\small 3,703} & {\small 6,805} & {\small 100} \\ \hline
{\small \#Nodes} & {\small 2,485} & {\small 2,110} & {\small 18,333} & {\small 2,449,029} \\ \hline
{\small \#Edges} &{\small 5,069} & {\small 3,668} & {\small 81,894} & {\small 61,859,140}\\ \hline
{\small \#Classes} & {\small 7} & {\small 6} & {\small 15} & {\small 47} \\ \hline
\end{tabular} \\
\label{dataset_stat} 
\end{table}

\subsubsection{Datasets}
We consider four benchmark graph datasets, i.e., Cora, Citeseer, Coauthor CS, and Amazon2M, that are used for node classification in previous works~\cite{kipf2017semi,shchur2018pitfalls,chiang2019cluster}. 
Cora and Citeseer~\cite{sen2008collective} are citation graphs, where a node indicates a document and an edge between two documents indicates one document cites the other. 
Each node's feature vector is the bag-of-words feature of the corresponding document and each document has a label.
Coauthor CS~\cite{shchur2018pitfalls} is a co-authorship graph based on the Microsoft Academic Graph
from the KDD Cup 2016 challenge.\footnote{\url{https://kddcup2016.azurewebsites.net/}} 
In this graph, a nodes means an author and an edge between two authors means the two authors co-authored a paper. Each node's feature vector indicates paper keywords of the author's papers, and node's label indicates the most active field of study of the author. 
Amazon2M is a large-scale graph dataset constructed by~\cite{chiang2019cluster}. Each node in the graph is a product, and each edge represents that two products are purchased together. Each node's feature vector is the bag-of-words feature extracted from the product descriptions. 
Statistics of these graph datasets are shown in Table~\ref{dataset_stat}. 
We note that these datasets have different graph properties, e.g., graph size,  averaged node degree, 
number of classes, etc.

\subsubsection{Compared learning methods}
We use two representative GNNs, i.e., GCN~\cite{kipf2017semi}\footnote{For Amazon2M, due to its large size, we adopt a memory- and time-efficient variant of GCN, called  Cluster-GCN~\cite{chiang2019cluster}, to train a local node classification model in each client.} and SGC~\cite{wu2019simplifying}, as the GraphSSC methods.
We compare our GraphFL with individual learning (IL) and  FL~\cite{mcmahan2016communication}.
\begin{itemize}
	\item {\bf Individual learning (IL).} In this method, we only have clients. Each client adopts the GraphSSC 
	method and trains a local node classification 
	model based on its few labeled nodes. Then, each learnt local model is used for classifying testing nodes and obtains a classification accuracy. We report the final classification accuracy as the averaged accuracy on all clients.

	\item {\bf Federated learning (FL)~\cite{mcmahan2016communication}.} In this method, we have both a center server and 
	several clients. Each client has a few labeled nodes and the server cannot access these labeled nodes. In each episode, the server initializes a global model and sends it to the selected participating clients; 
	each selected client adopts the GraphSS method to train a local model using its few labeled nodes and the global model; and the server updates the global model by averaging the local model parameters of the selected clients. After several episodes, the server uses the final global model to classify testing nodes.

\end{itemize}

\subsubsection{Training set and testing set.}
For experiments aiming to address the non-IID issue in graph data, we randomly sample 80 nodes from each class in each dataset to form the training set and evenly distribute these labeled nodes to all clients (i.e., 50 clients in our experiments). We note that each client will be assigned a very few labeled nodes in total 
(between 9 and 70), which implies that labeled nodes across clients could be 
highly non-IID.
The training set is further splitted into two halves, where the first part is either used as the training nodes in compared methods or as the support nodes in our method; and the second part is either used for hyperparameter tuning in compared methods or as the query nodes in our method.  
Moreover, consider different graph sizes of our datasets, we randomly select 1,000 nodes in the Cora and Citeseer graphs, and 10,000 nodes in the Coauthor CS and Amazon2M graphs as the testing set. 
We sample the training set and testing set 5 times and report the average classification accuracies as the final result.

For experiments aiming to classify testing nodes with new label domains, we split the nodes with $K$ classes in each graph dataset into two separate sets. The first set contains nodes from the first $K-K_0$ classes and the second set contains nodes from the remaining $K_0$ classes. 
We use different $K_0$ in different datasets considering their different number of classes. 
We sample nodes for each client from the first set to form the training set, and sample nodes from the second set to form the testing set. 
In doing so, nodes in testing set and training set have distinct label domains. 
Specifically, we randomly sample $K_0$ classes per client from the first set, with each class sampling 10 nodes to form the training set. The training set are further splitted into the support nodes and the query nodes in our method; or the training nodes and the validation nodes in the compared FL method. 
We also sample $K_0$ classes per client from the second set, with each class sampling the same number of nodes as the support nodes for fast adaptation/finetuning and sampling 20 nodes as the testing nodes.

\subsubsection{Parameter settings.}
In our method, we set $I=50$ clients in total, number of episodes $T=50$, steps of local gradient descent $T_e=15$. 
We randomly intialize the global model $\theta_0$, 
In experiments that classify testing nodes with new label domains, we set $K_0=2$ in Cora and Citeseer and $K_0=3$ in Coauthor CS and Amazon2M. 
There are three key parameters that could affect the performance of our method: 
fraction of participating clients per episode, number of labeled nodes in the training set, and fraction of overlapping between client graphs. 
By default, we assume 20\% participating clients per episode, 80 labeled nodes per class as the training set, and all clients have the complete graph. 
When studying the impact of each parameter, we fix the remaining ones to be their default values.

\subsection{Node classification results with non-IID graph data}
In this section, we evaluate our method and compare it with  individual  learning  (IL), and FL for federated GraphSSC 
with non-IID graph data across clients.

\begin{figure*}[!t]
\center
\subfloat[Cora]{\includegraphics[width=0.248\textwidth]{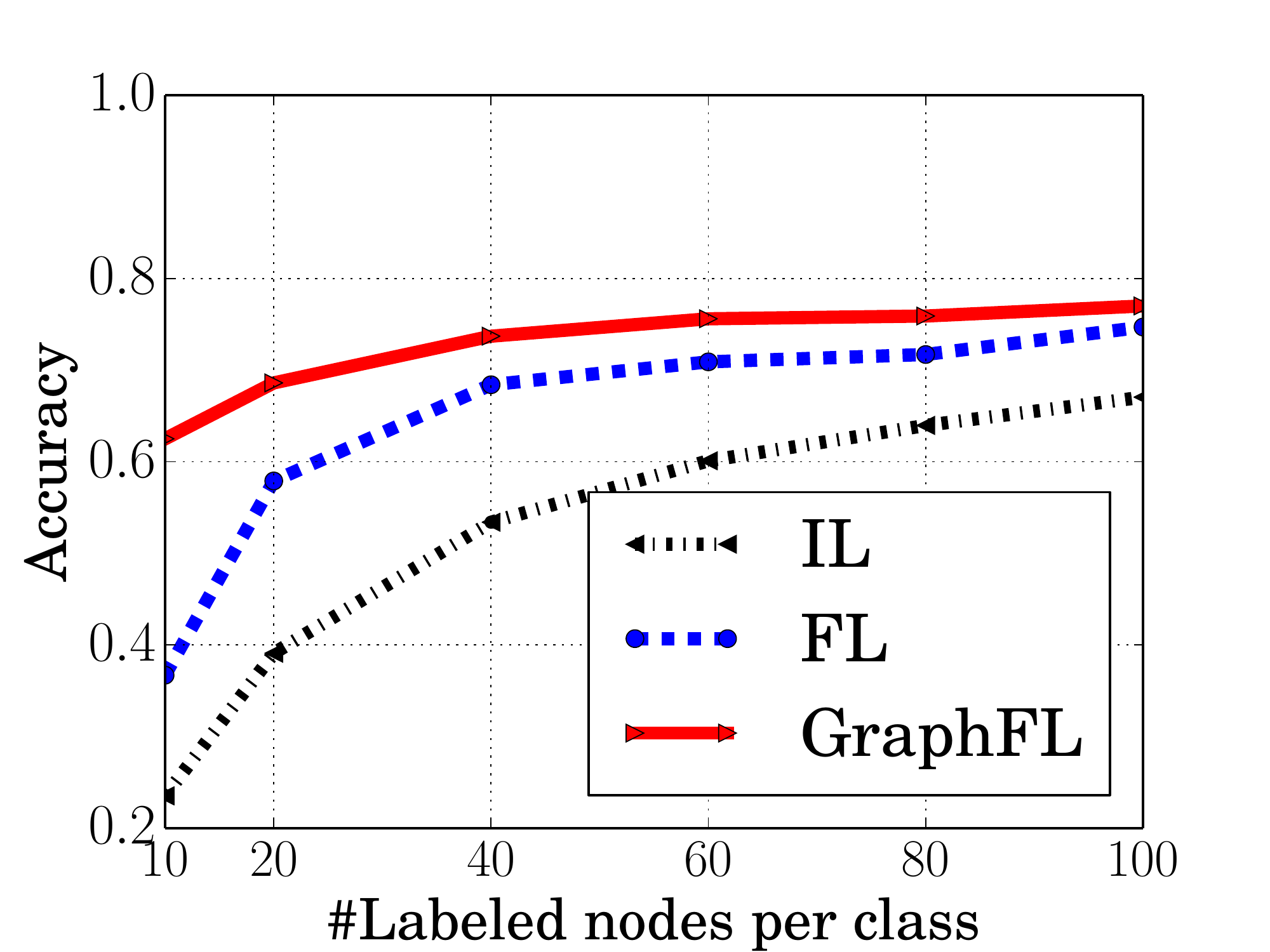}}
\subfloat[Citeseer]{\includegraphics[width=0.248\textwidth]{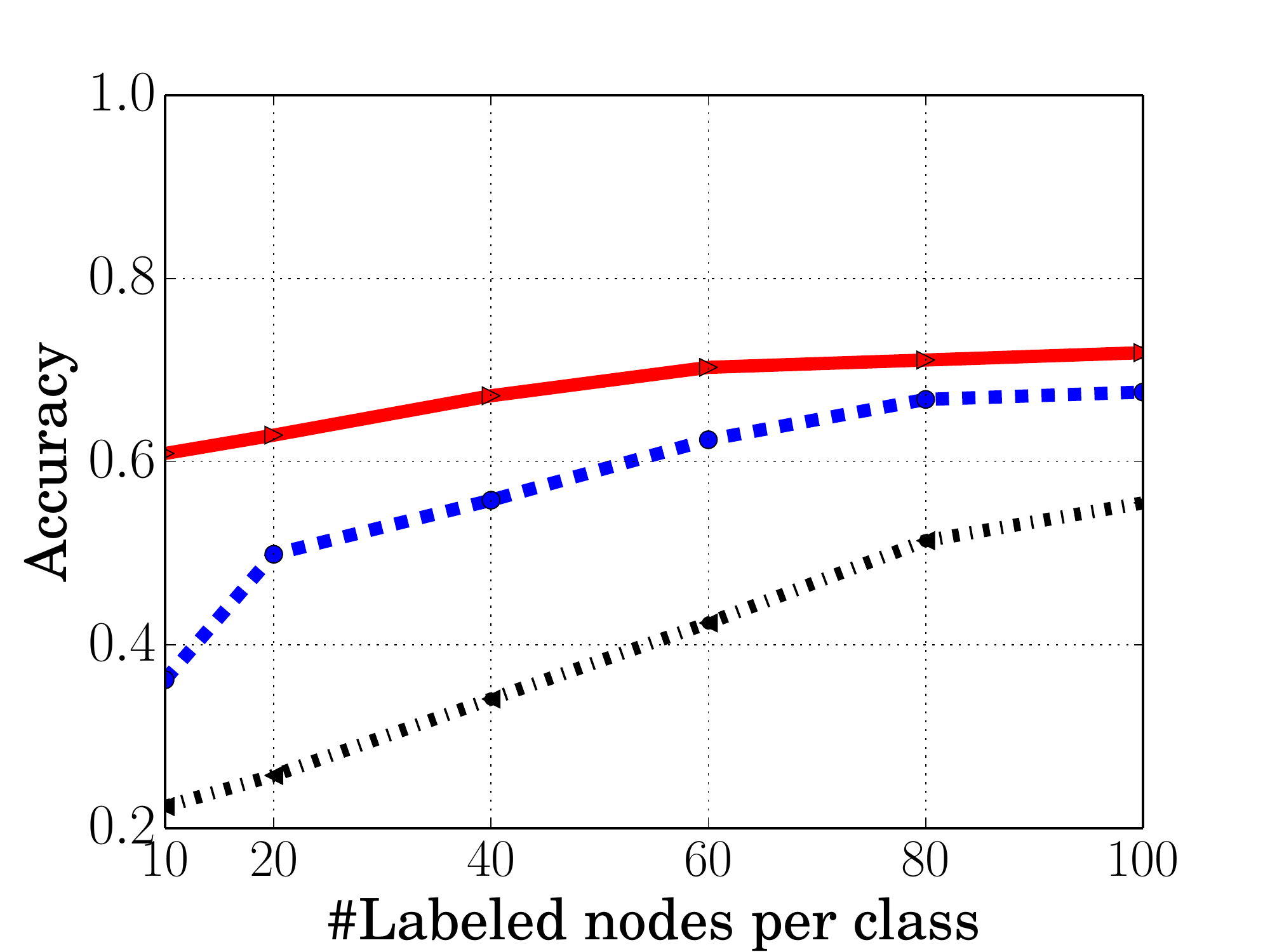}}
\subfloat[Coauthor CS]{\includegraphics[width=0.248\textwidth]{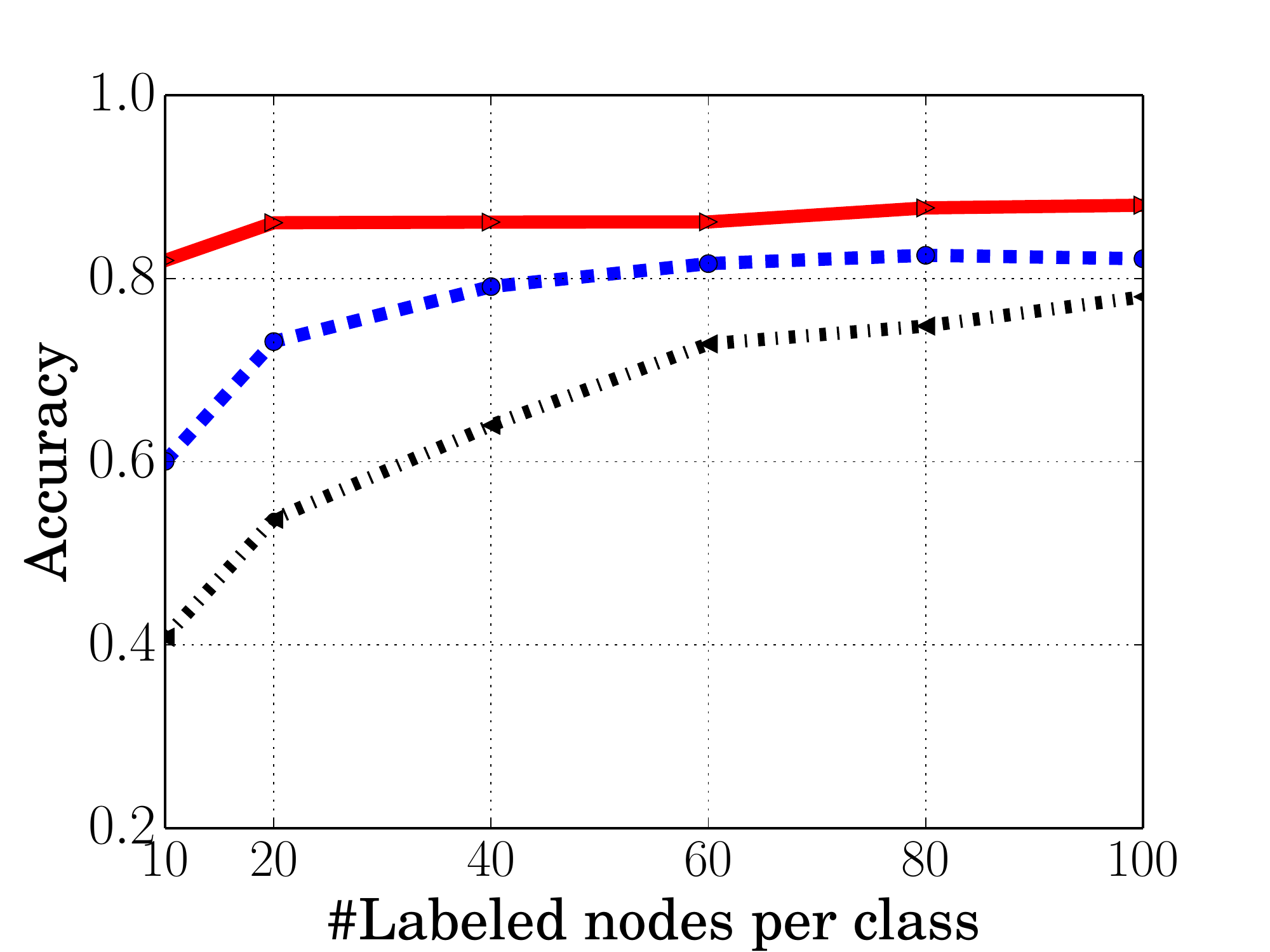}}
\subfloat[Amazon2M]{\includegraphics[width=0.248\textwidth]{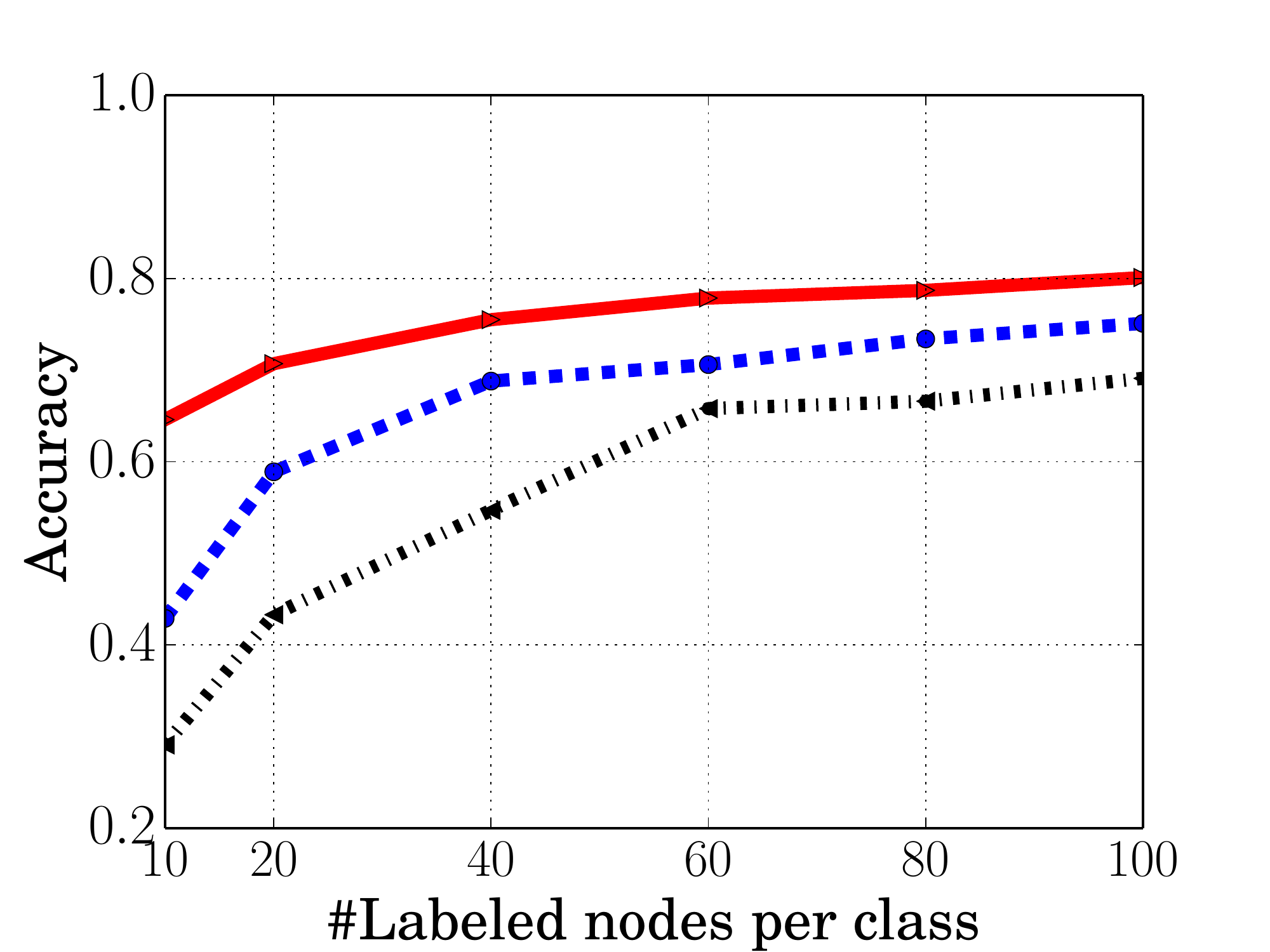}}
\caption{Node classification accuracy of GCN using the compared learning methods vs. number of labeled nodes per class.}
\label{impact_nodes_GCN}
\end{figure*}

\begin{figure*}[!t]
\center
\subfloat[Cora]{\includegraphics[width=0.248\textwidth]{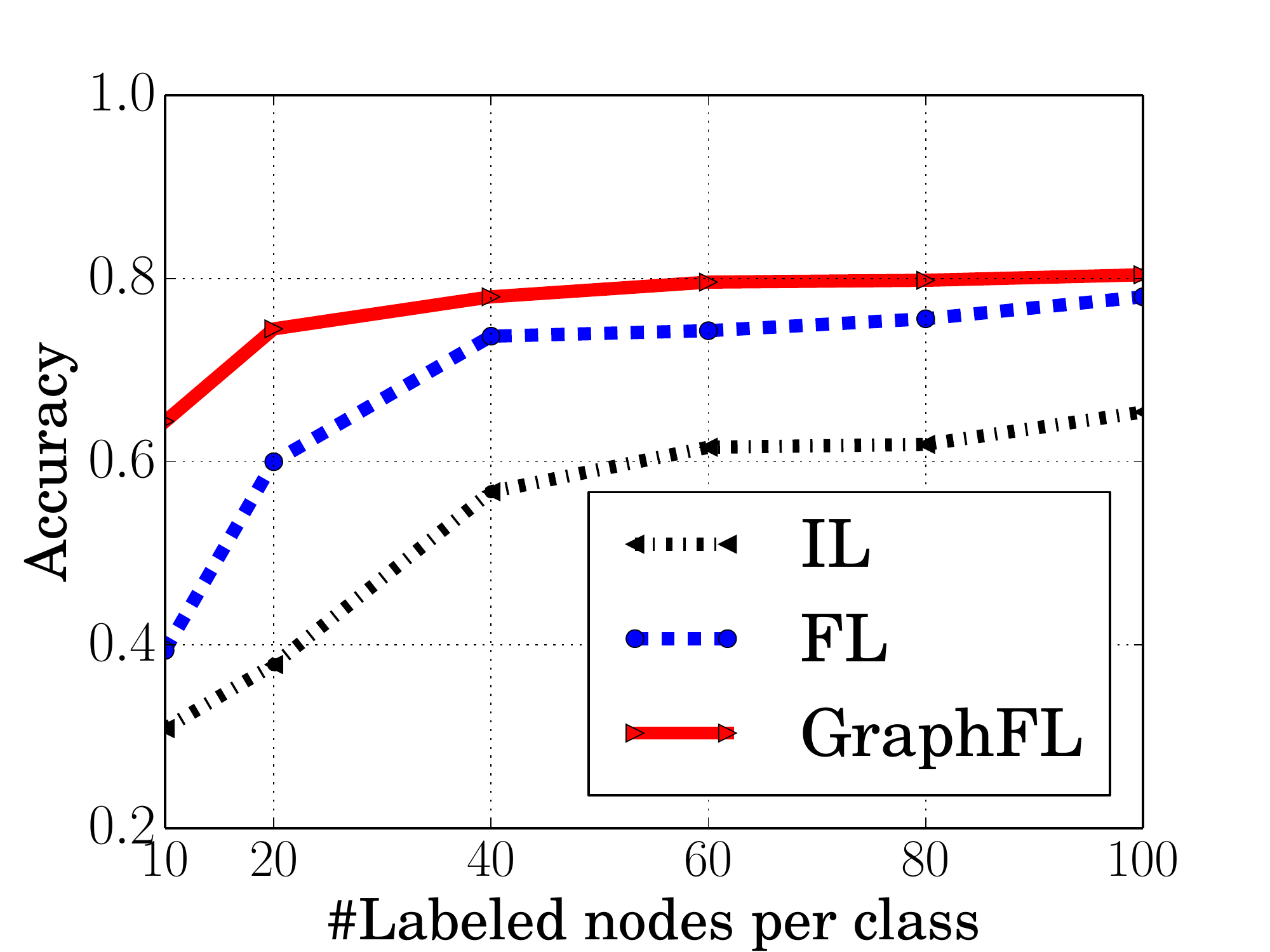}}
\subfloat[Citeseer]{\includegraphics[width=0.248\textwidth]{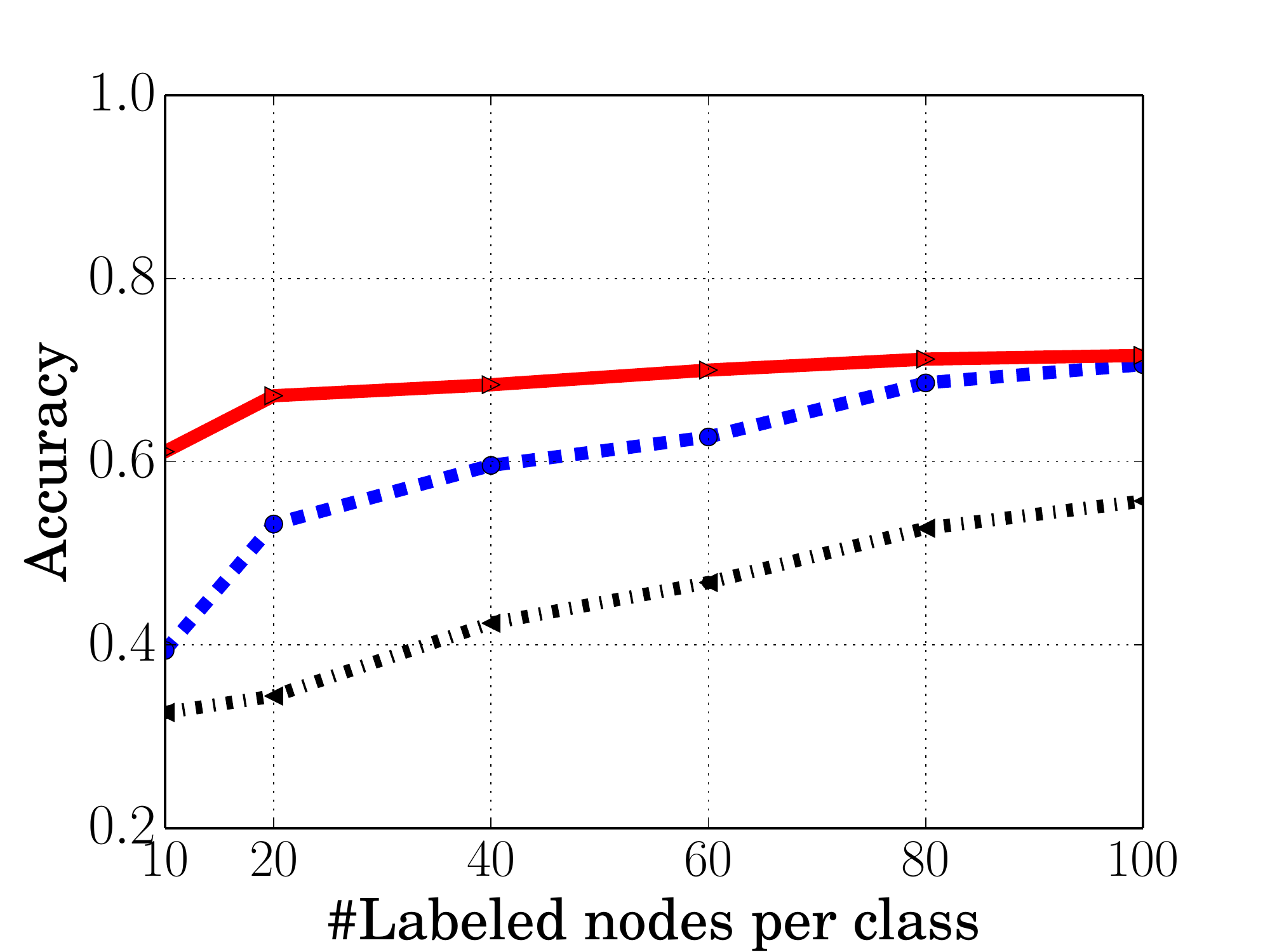}}
\subfloat[Coauthor CS]{\includegraphics[width=0.248\textwidth]{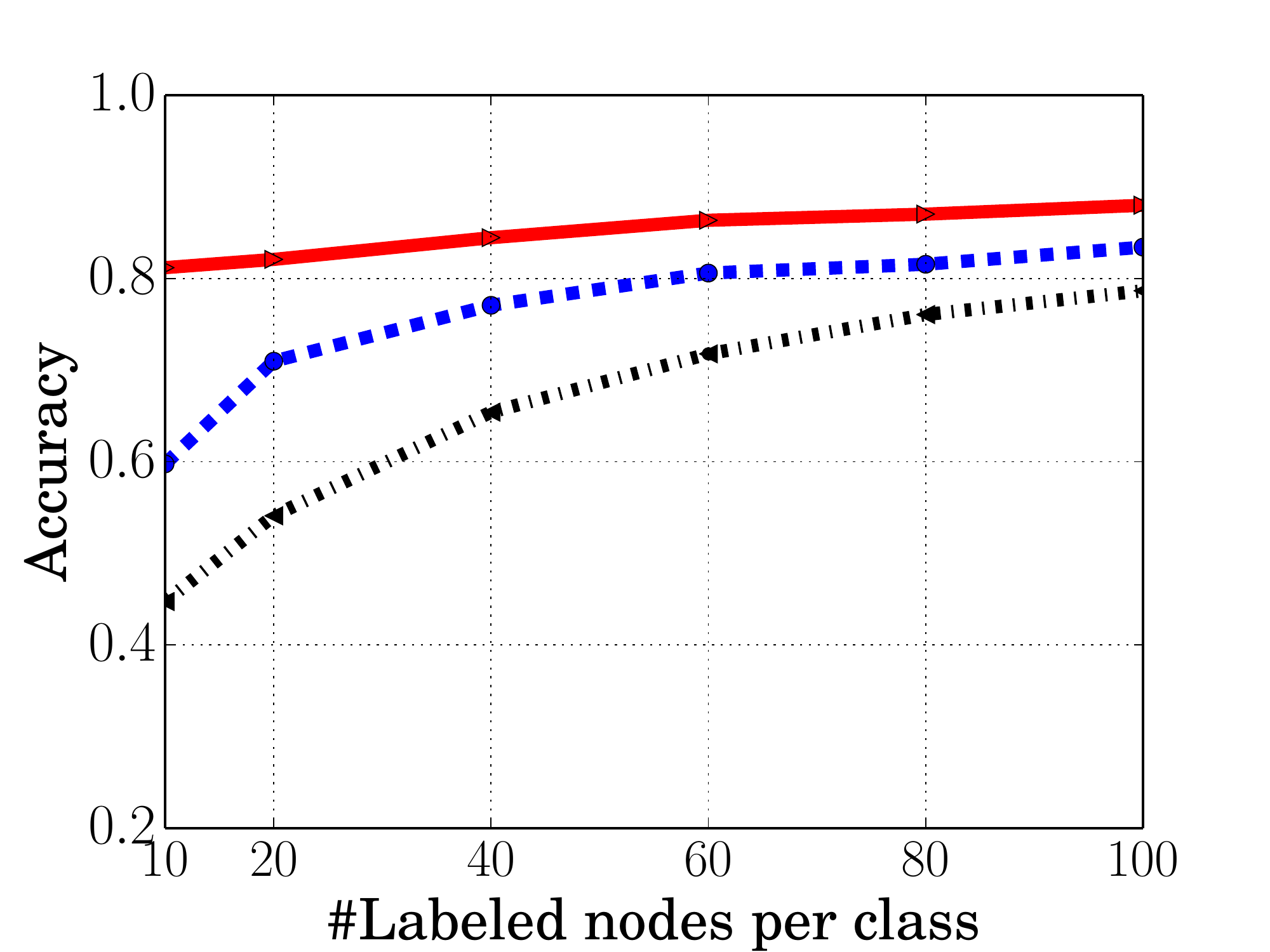}}
\subfloat[Amazon2M]{\includegraphics[width=0.248\textwidth]{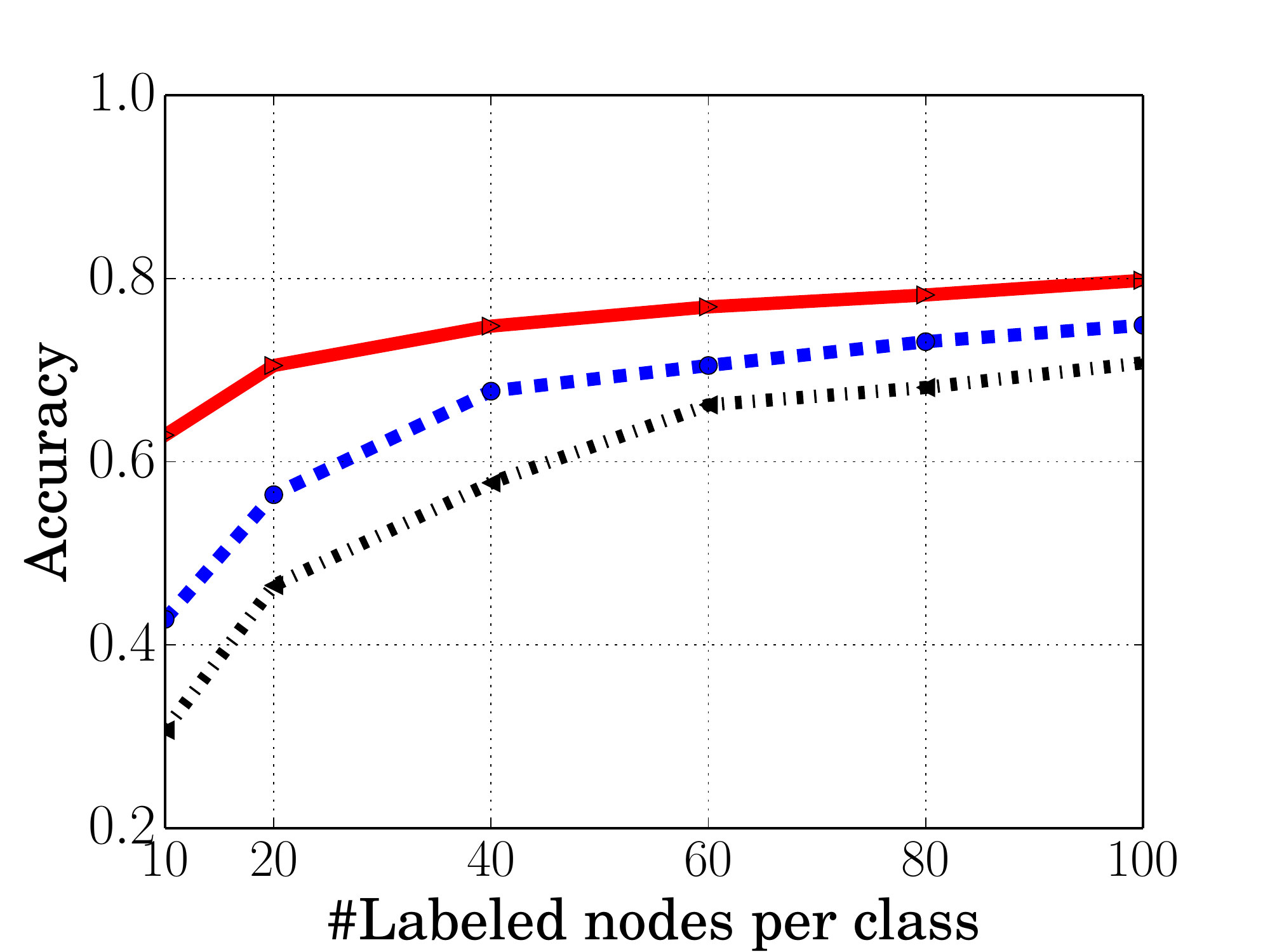}}
\caption{Node classification accuracy of SGC using the compared learning methods vs. number of labeled nodes per class.}
\label{impact_nodes_SGC}
\end{figure*}

\subsubsection{Impact of the number of labeled nodes per class.} 
Figure~\ref{impact_nodes_GCN} and Figure~\ref{impact_nodes_SGC} respectively show the node classification accuracy of GCN and SGC using the compared learning methods 
vs. number of labeled nodes per class on the four datasets. 
We have the following observations. 
First, IL performs the worst in all cases. This is because each client can only leverage its own labeled nodes in IL, while the other two learning methods can leverage labeled nodes' information from all the other clients. 
Second, our GraphFL consistently outperforms FL. This is because GraphFL leverages the training scheme of MAML, which can better handle the non-IID issue in graph data than existing FL. 
We note that, when the number of labeled nodes per class is smaller (e.g., 10 labeled nodes per class),
which indicates more serious non-IID, the accuracy gap between GraphFL and FL is larger, i.e., $\sim$20\%. 
In the following, we mainly compare FL and GraphFL for conciseness.

\subsubsection{Impact of the fraction of participating clients per episode.}
\begin{figure*}[!t]
\center
\subfloat[Cora]{\includegraphics[width=0.248\textwidth]{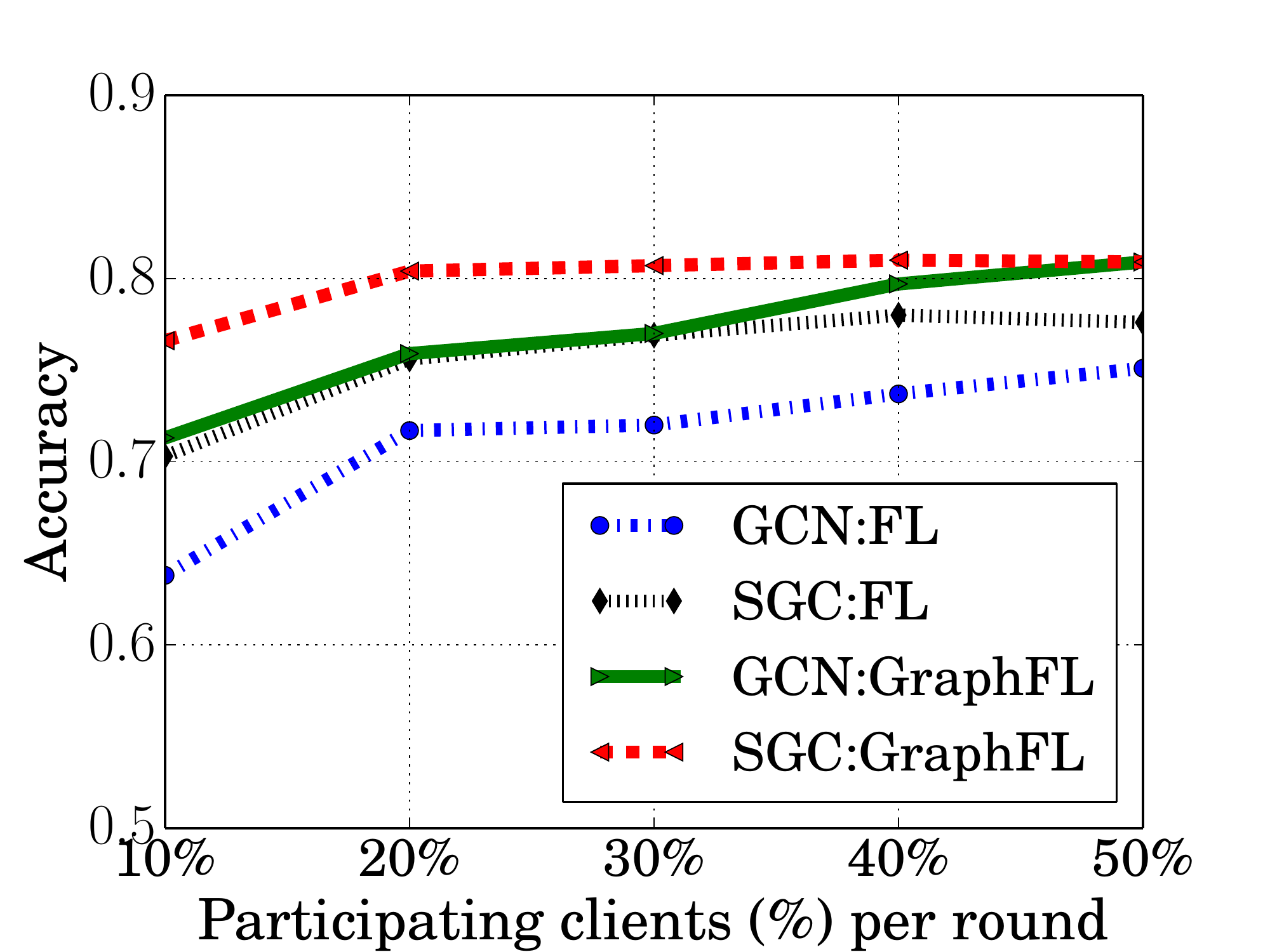}}
\subfloat[Citeseer]{\includegraphics[width=0.248\textwidth]{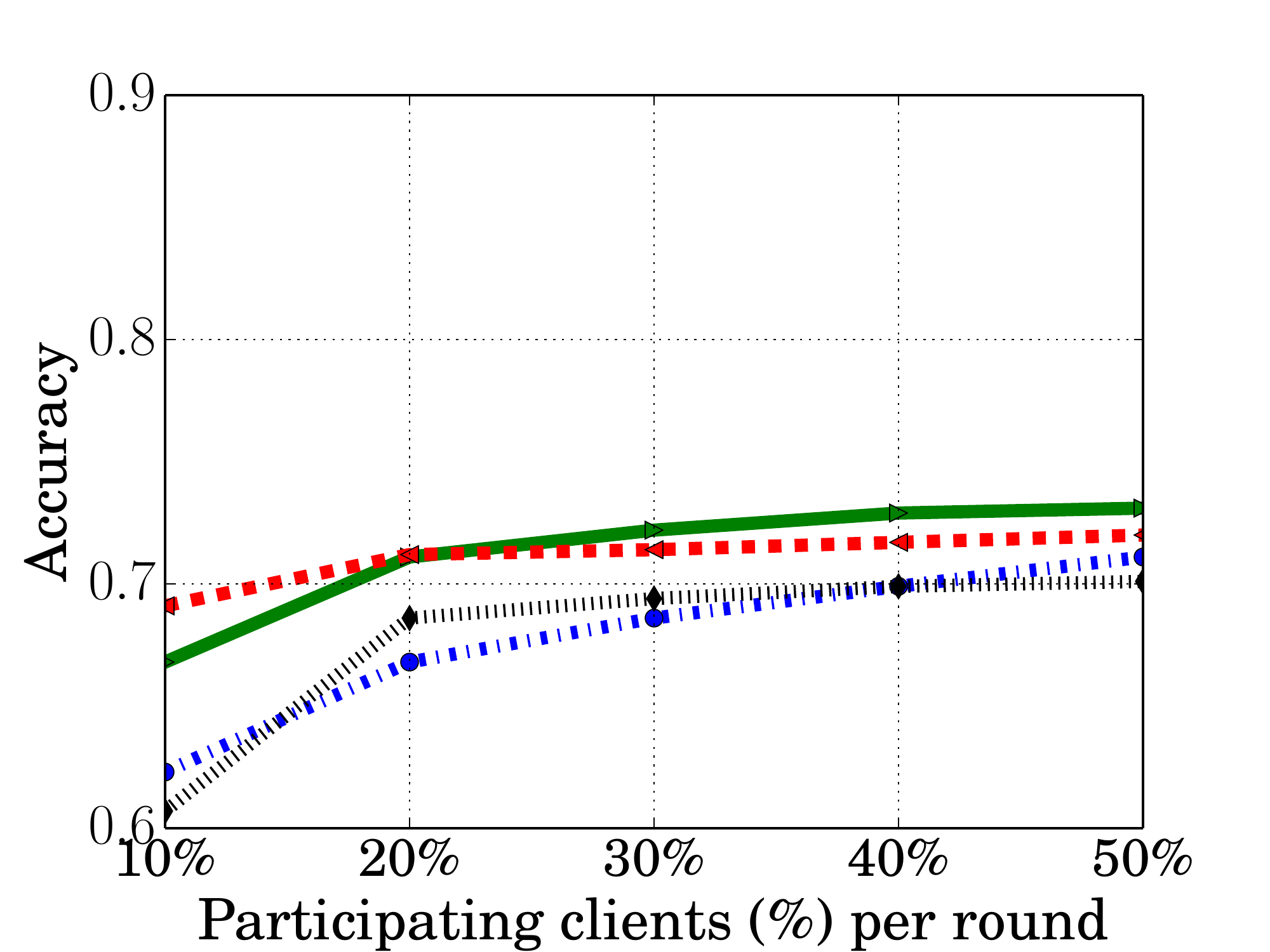}}
\subfloat[Coauthor CS]{\includegraphics[width=0.248\textwidth]{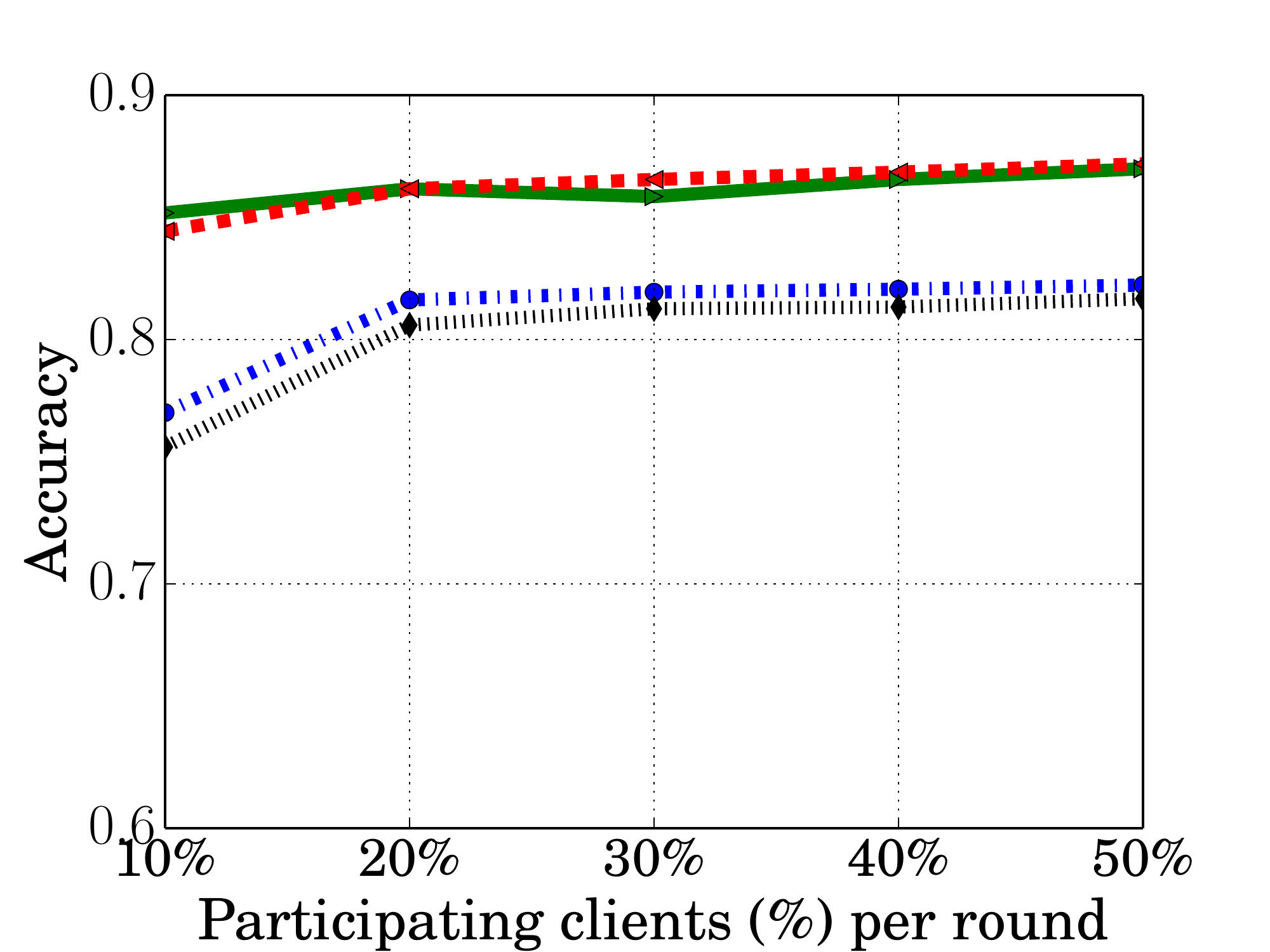}}
\subfloat[Amazon2M]{\includegraphics[width=0.248\textwidth]{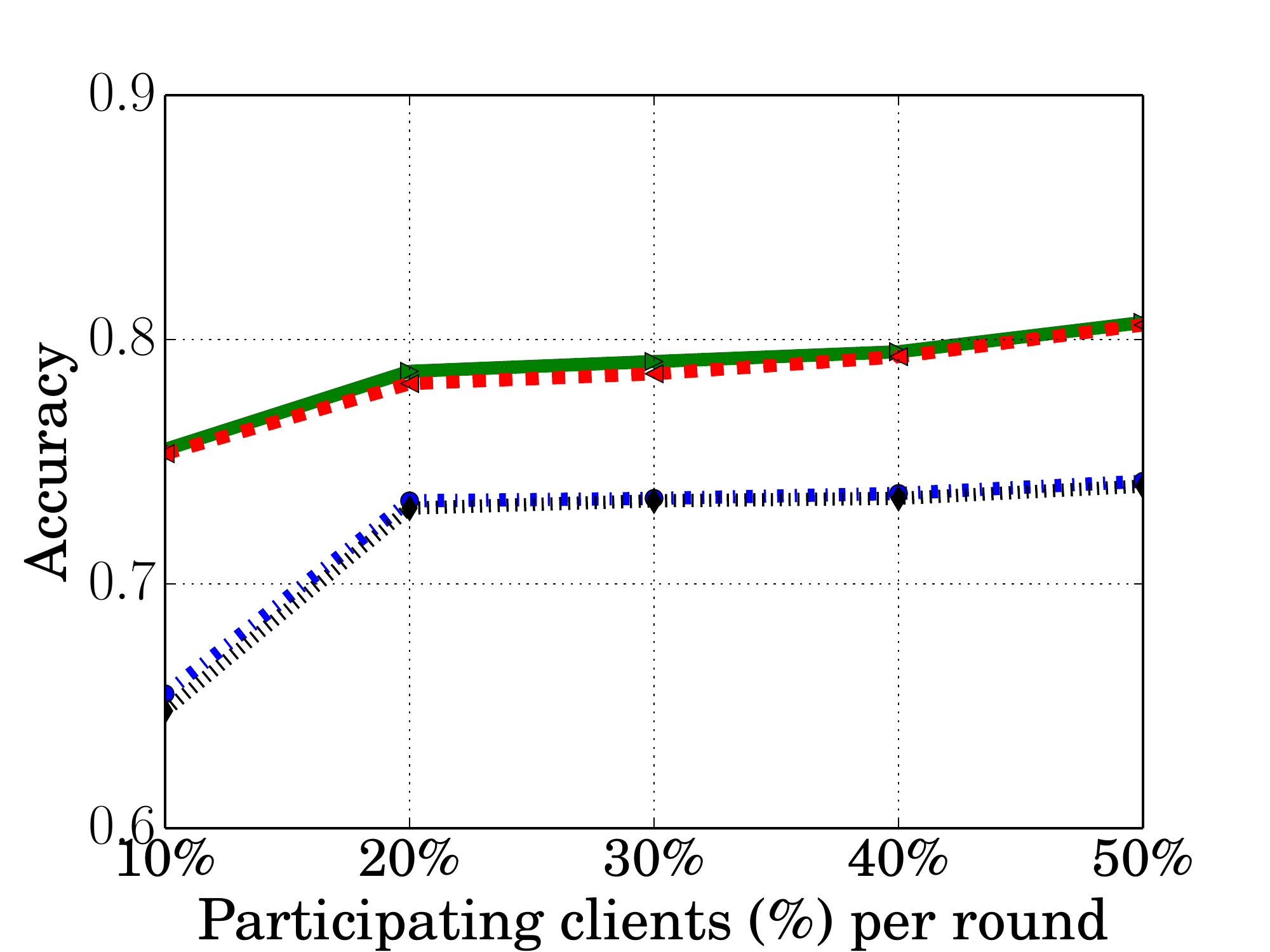}}
\caption{Node classification accuracy of GCN and SGC using FL and GraphFL vs. fraction of participating clients per round.}
\label{impact_clients}
\end{figure*}

Figure~\ref{impact_clients} shows the node classification accuracy of GCN and SGC using FL and GraphFL vs. fraction of participating clients per episode on the four datasets. 
First, both FL and our GraphFL can assist GCN and SGC to achieve higher accuracies, as the fraction of participating clients becomes larger. This is because both methods can leverage more labeled nodes when a larger number of clients participates in the training. 
Second, our GraphFL outperforms FL in all datasets. Moreover, our GraphFL has a larger performance gain than FL, when the fraction of participating clients is smaller. One possible reason is that a smaller number of participating clients results in higher non-IID distribution across clients' labeled nodes.

\begin{figure}[!t]
\center
\subfloat[Coauthor CS]{\includegraphics[width=0.45\textwidth]{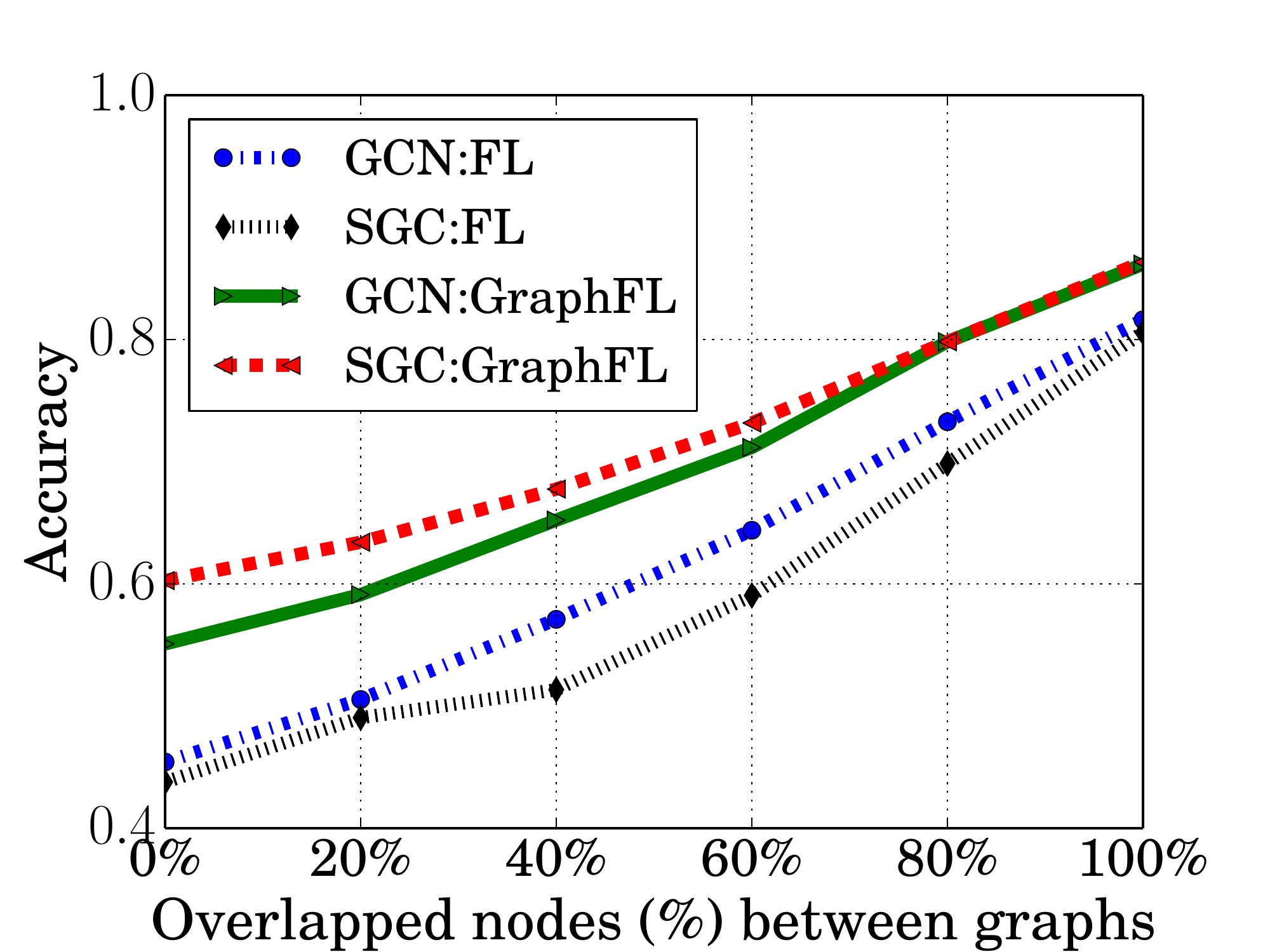}}
\subfloat[Amazon2M]{\includegraphics[width=0.45\textwidth]{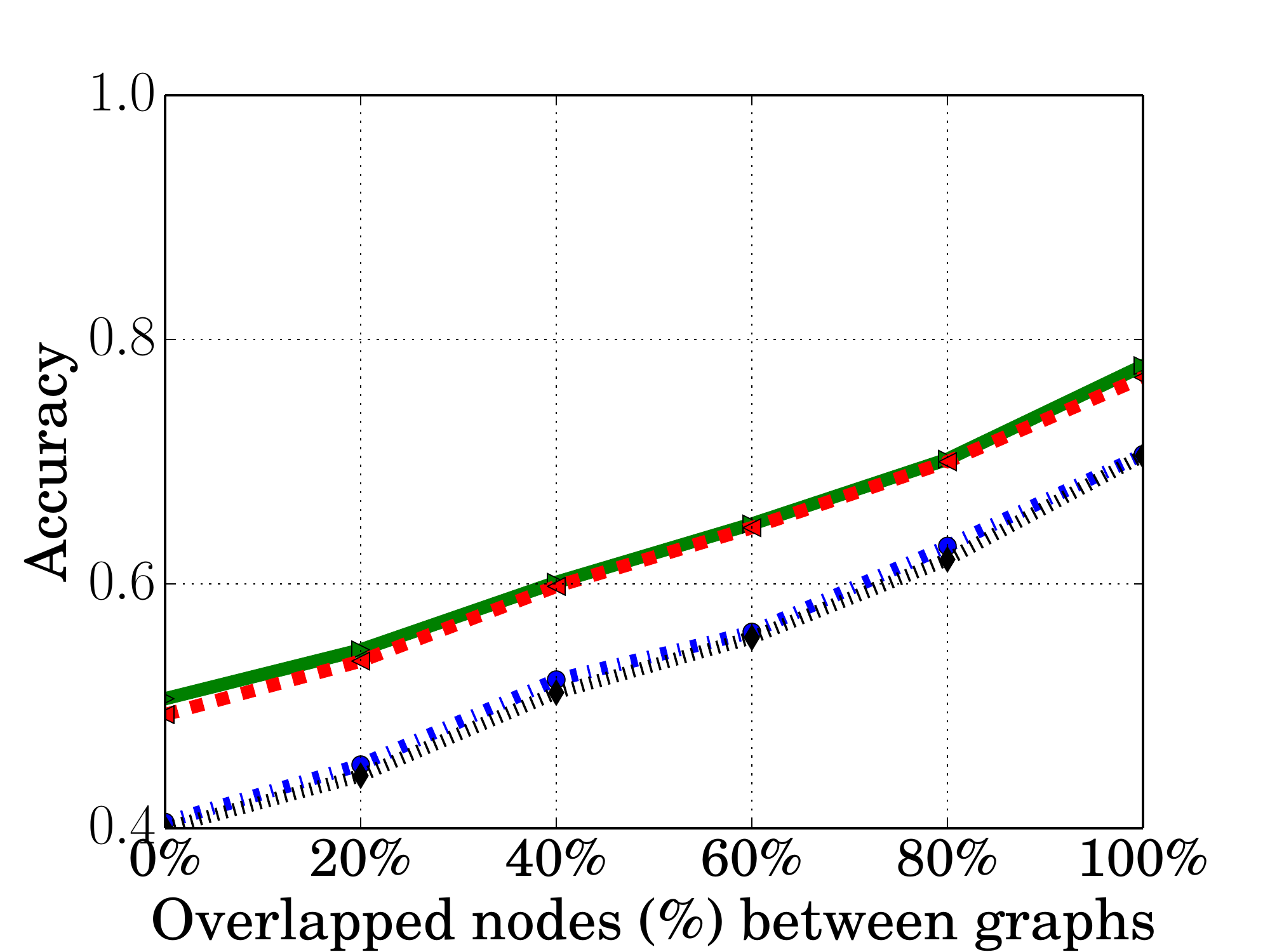}}
\caption{Node classification accuracy of GCN and SGC using FL and GraphFL vs. fraction of overlapped nodes between graphs.}
\label{impact_graphsize}
\end{figure}

\subsubsection{Impact of the fraction of overlapped nodes between client graphs.}
In this experiment, we simulate real-world scenarios where each client is assumed to have a partial graph of a real large graph. 
To simplify the simulation, we use the fraction of overlapped nodes between client graphs as a metric. 
Specifically, given a large graph and a fraction $\gamma$, we evenly distribute the total nodes to all clients such that any two clients with neighboring indexes have a fraction $\gamma$ of overlapped nodes. 
Then, each client graph consists of the distributed nodes and the associated edges connecting these nodes. We ignore the edges that are also associated with nodes in other clients. 
Consider that the graph size of Cora and Citesser is rather small, we only conduct experiments on the Coauthor CS and Amazon2M datasets for simplicity. 

Figure~\ref{impact_graphsize} shows the node classification accuracy of GCN and SGC using FL and our GraphFL vs. fraction of overlapped nodes between clients. 
We have the following observations. 
First, as the fraction of overlapped nodes between graphs increases, node classification accuracy of both FL and GraphFL on GCN and SGC also increases. 
Second, our GraphFL consistently outperforms FL. 
Moreover, when there are no overlaps between client graphs, our GraphFL has the largest performance gain over FL. This is because in this case graph data across clients are the most non-IID.

\begin{table*}[tbh]
\caption{Accuracy of GCN and SGC using FL and GraphSSC with new label domains vs. \#labeled nodes per class.}
\centering
\addtolength{\tabcolsep}{-4pt}
\begin{tabular}{|c|c|c|c|c|c|c|c|c|c|c|c|c|}
\hline
{\bf GCN} & \multicolumn{3}{c|}{\bf Cora} & \multicolumn{3}{c|}{\bf Citeseer} & \multicolumn{3}{c|}{\bf Coauthor CS} & \multicolumn{3}{c|}{\bf Amazon2M} \\ \hline
 {\bf \#Labels} & 2     &   6    &   10    &  2     &   6    &  10     &   2    &   6    &   10    &   2    &   6    &   10    \\ \hline
 {\bf FL+TL} &   0.527    &  0.623     &   0.667    &    0.500   &  0.503     &   0.527    &   0.664	    &  0.667	     &   0.711	    &   0.614	    &  0.626     &  0.637     \\ \hline
 {\bf GraphFL}		&   {\bf 0.667}    &  {\bf 0.767}     &  {\bf 0.843}     &   {\bf 0.620}    &   {\bf 0.630}    &  {\bf 0.670}    &   {\bf 0.774}    &   {\bf 0.835}	    &    {\bf 0.889}   &   {\bf 0.699}	   &  {\bf 0.706}     &   {\bf 0.764}	    \\ \hline \hline

 {\bf SGC} & \multicolumn{3}{c|}{\bf Cora} & \multicolumn{3}{c|}{\bf Citeseer} & \multicolumn{3}{c|}{\bf Coauthor CS} & \multicolumn{3}{c|}{\bf Amazon2M} \\ \hline
 {\bf \#Labels } & 2     &   6    &   10    &  2     &   6    &  10     &   2    &   6    &   10    &   2    &   6    &   10    \\ \hline
 {\bf FL+TL} &   0.500    &  0.517     &   0.543    &    0.453   &  0.483     &   0.494    &   0.607	    &  0.647	     &   0.687	    &   0.601	    &  0.614     &  0.626     \\ \hline
 {\bf GraphFL}		&   {\bf 0.647}    &  {\bf 0.710}     &  {\bf 0.773}     &   {\bf 0.583}    &   {\bf 0.603}    &  {\bf 0.653}    &   {\bf 0.740}    &   {\bf 0.806}	    &    {\bf 0.862}   &   {\bf 0.697}	    &  {\bf 0.702}     &  {\bf 0.757}	    \\ \hline
\end{tabular}
\label{impact_clients_cross} 
\end{table*}

\subsection{Node classification results with new label domains}
\label{ood_eval}
In this section, we evaluate our GraphFL and compare it with FL for federated semi-supervised node classification with testing nodes having new label domains. 
As existing FL cannot handle nodes with new label domains, we adopt the transfer learning (TL) solution as described above. 
We only show the results on using different number of labeled nodes per class. 
Note that we also find similar observations as in aforementioned results on the impact of the fraction of participating clients and the fraction of overlapped nodes between client graphs.

Table~\ref{impact_clients_cross} shows the node classification accuracy of GCN and SGC using FL and our GraphFL with new label domains vs. number of labeled nodes per class on the four graph datasets. 
We observe that our GraphFL method consistently outperforms FL, with at least a 10\% higher accuracy in almost all cases. This indicates that transfer learning is not effective enough in handling testing nodes with new label domains for federated GraphSSC. In addition, when the number of labeled nodes per class is bigger, the performance gain of our GraphSSL over FL + TL is larger.

\subsection{Node classification results with self-training}
In this experiment, we study the effectiveness of adopting self-training to further enhance GraphFL's  performance. 

\begin{figure}[t]
\vspace{-4mm}
\center
\subfloat[GCN]{\includegraphics[width=0.4\textwidth]{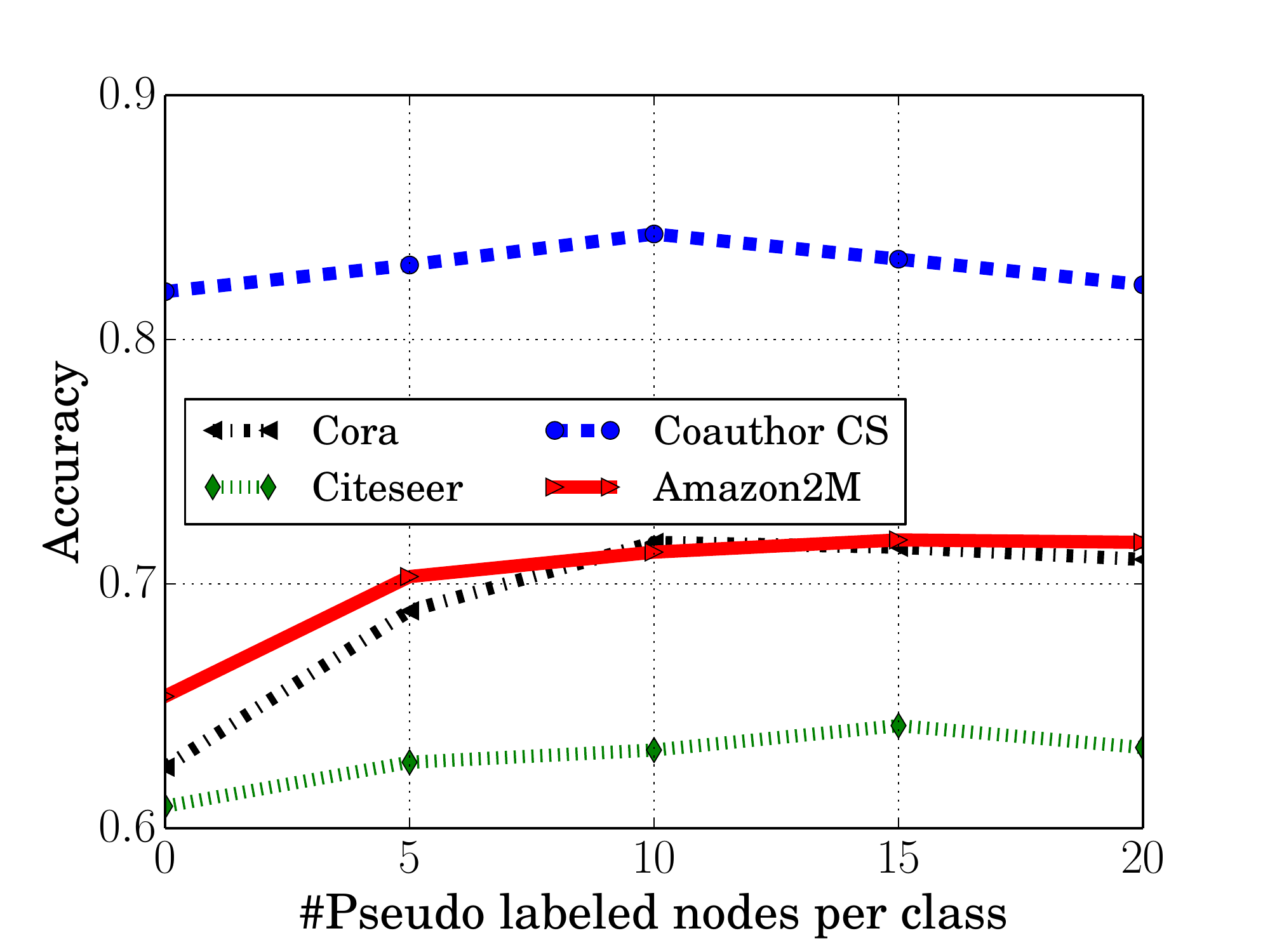}}
\subfloat[SGC]{\includegraphics[width=0.4\textwidth]{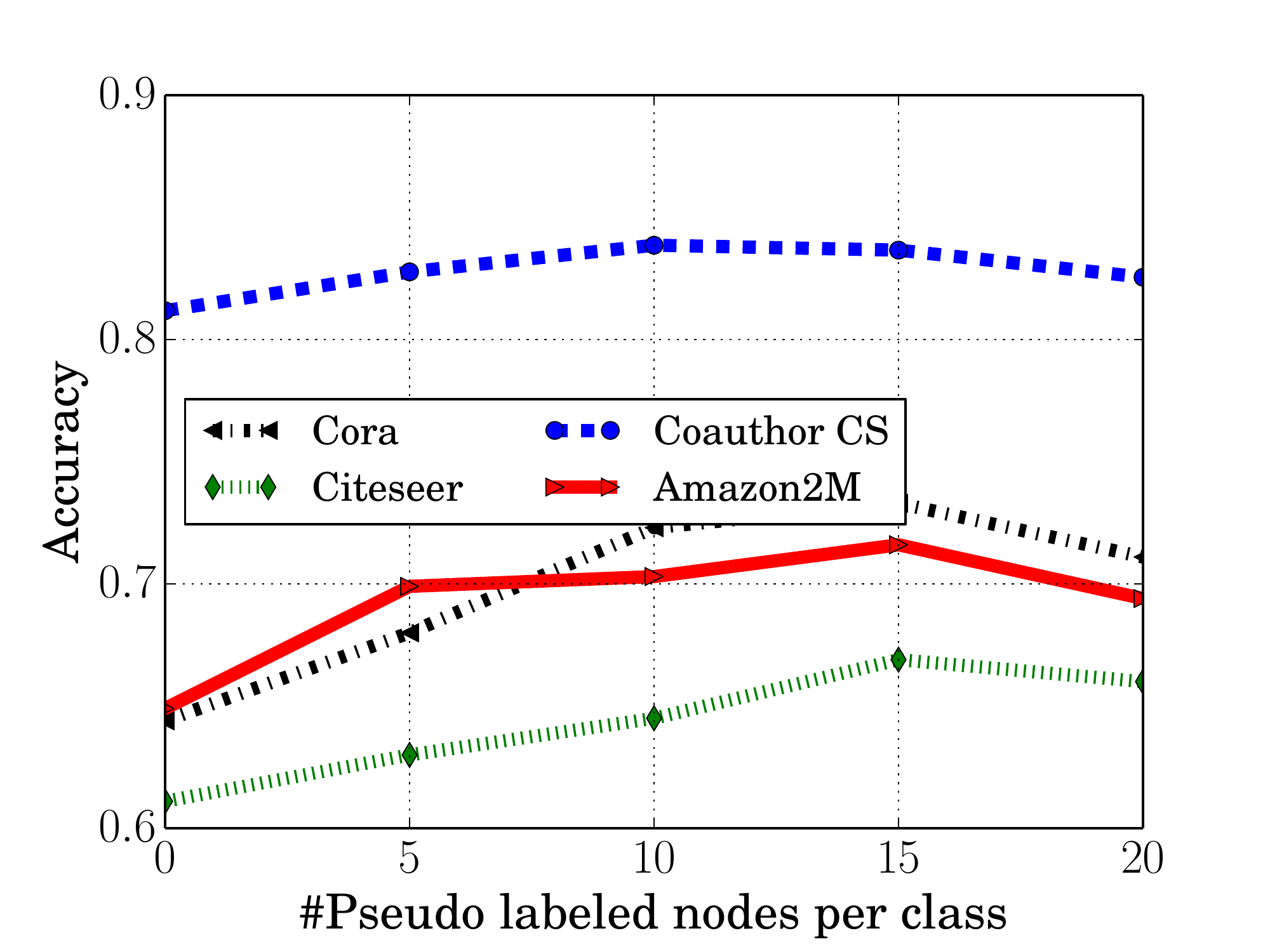}}
\caption{Node classification accuracy of (a) GCN and (b) SGC using GraphFL vs. \#pseudo labeled nodes per class.}
\label{impact_self_train}
\end{figure}

Figure~\ref{impact_self_train} shows the node classification accuracy of GCN and SGC using our GraphFL vs. number of pseudo labeled nodes per class generated via self-training (Due to space limitation, we do not show results with new label domains. However, we have similar observations. 
First, our method obtains a higher accuracy on all datasets when it uses the pseudo labeled nodes for training. 
This demonstrates that the pseudo labeled nodes are indeed beneficial to the federated training and thus can further enhance the performance of our method for semi-supervised node classification. 
One reason is that most of the predicted labels of pseudo labeled nodes match these nodes' true labels. 
Second, as the number of pseudo labeled nodes per class increases, node classification accuracy first increases and then decreases. 
This is because, when the number of pseudo labeled nodes is relatively small, most of these nodes' labels are correctly predicted via self-training. However, as the number becomes larger, wrong predicted labels of the pseudo labeled nodes also becomes larger. 
Table~\ref{purity} also shows the fraction of the pseudo labeled nodes whose predicted labels are correct via GCN self-training on the four datasets. We note that these numbers justify our above claims.

\begin{table}[t]\renewcommand{\arraystretch}{1.2}
\caption{Fraction of the pseudo labeled nodes with correct predictions via GCN self-training.}
\centering
\begin{tabular}{|c|c|c|c|c|}
\hline
{\bf \#Pseudo labeled nodes} & {\bf 5} & {\bf 10} & {\bf 15} & {\bf 20} \\ \hline
{\bf Cora} & 90\% & 85\% & 81\% & 69\% \\ \hline
{\bf Citeseer} & 90\% & 80\% & 72\% &  64\% \\ \hline
{\bf Coauthor CS} & 95\% &  90\% & 83\% & 74\% \\ \hline
{\bf Amazon2M} & 83\% & 76\% & 73\% & 69\% \\ \hline
\end{tabular}
\label{purity}
\end{table}

\subsection{Summary}
\begin{itemize}
	\item Our method significantly outperforms conventional FL for federated GraphSSC in terms of handling non-IID graph data and generalizing to testing nodes with new label domains. 
	\item GraphFL can be further enhanced when unlabeled nodes are leveraged via self-training. 
	\item Increasing the number of training nodes, the fraction of participating clients, the graph size in clients, or 
	leveraging unlabeled nodes all 
	can all enhance the performance of federated node classification on graphs. 
\end{itemize}

\section{Conclusion}
\label{conclusion}

We study federated semi-supervised node classification on graphs and propose the first federated learning framework called GraphFL. 
Graph-based semi-supervised node classification has its unique challenges when studied under the FL setting, such as 
1) graph data across clients are possibly highly non-IID; 2) testing nodes and training nodes could have different label domains; 3) client graphs have a substantial number of unlabeled nodes. 
We propose two MAML-inspired GraphFL methods to respectively address 1) and 2). 
Moreover, we design a self-training method to address 3). 
We evaluate our GraphFL methods on two representative graph neural networks for federated semi-supervised node classification. Our results on multiple graph datasets demonstrate that our methods significantly outperform the compared baselines and our methods with self-training can obtain better performance.

\bibliographystyle{plain}
\bibliography{refs}

\end{document}